\documentclass{article}

\title{\huge {Hyena Hierarchy:}\\Towards Larger Convolutional Language Models}
 \author{Michael Poli\footnote{Equal contribution. $\dagger$ Equal senior authorship. $^1$Stanford University. $^2$Mila and Universit\'e de Montr\'eal.}~$^{,1}$, Stefano Massaroli$^{*,2}$, Eric Nguyen$^{1,*}$, \\ Daniel Y. Fu$^1$, Tri Dao$^1$, Stephen Baccus$^1$, \\ Yoshua Bengio$^2$, Stefano Ermon$^{1,\dagger}$, Christopher R\'e$^{1,\dagger}$
 }
 
\date{\small{\footnotesize\sf Version}: submitted draft, {\footnotesize\sf  Last Compiled}: \today}
\usepackage{lipsum}
\usepackage[utf8]{inputenc}
\usepackage[T1]{fontenc}    
\usepackage[bookmarks=true, colorlinks=true,linkcolor=blue!70,citecolor=green!80!black]{hyperref}
\usepackage[round]{natbib}
\usepackage{wrapfig}
\usepackage{booktabs}  
\usepackage{longtable}
\usepackage{ducksay}
\usepackage{multicol}
\usepackage{etaremune}
\usepackage{afterpage}
\usepackage{capt-of}
\usepackage[table, x11names]{xcolor}
\usepackage{decorule}
\usepackage{scalerel,xparse}
\usepackage{enumitem} 
\usepackage[top=25mm, bottom=25mm, left=25mm, right=25mm,
            foot=5mm, marginparsep=0mm]{geometry}

\usepackage{chngcntr}
\counterwithin{figure}{section} 
\counterwithin{table}{section}

\usepackage{graphicx}
\usepackage{tikz}
\usepackage{tikz-cd}
\usepackage{hf-tikz}
\usepackage{pgfplots} 
\pgfplotsset{compat=newest} 
\pgfplotsset{
        table/search path={figures/drawings},
    }
\usetikzlibrary{fadings}
\usetikzlibrary{shapes, arrows, fit, backgrounds, arrows.meta}

\usetikzlibrary{matrix}
\usetikzlibrary{shadows.blur}
\usetikzlibrary{patterns, tikzmark}
\usetikzlibrary{decorations.pathreplacing, calc, decorations.markings,}
\usetikzlibrary{positioning}
\usetikzlibrary{shapes.geometric}
\pgfdeclareplotmark{mystar}{
    \node[star,star point ratio=2.25,minimum size=5pt,
          inner sep=0pt,draw=black,solid,fill=green] {};
}

\usepgfplotslibrary{groupplots}
\usepgfplotslibrary{patchplots}

\definecolor{bg}{gray}{0.97}
\definecolor{olive}{rgb}{0.6, 0.6, 0.2}
\definecolor{sand}{rgb}{0.8666666666666667, 0.8, 0.4666666666666667}
\definecolor{wine}{rgb}{0.5333333333333333, 0.13333333333333333, 0.3333333333333333}
\definecolor{deblue}{RGB}{11,132,147}
\definecolor{ocra}{RGB}{204, 119, 34}

\def\ocra{\color{ocra}}

\usepackage{lettrine}
\usepackage{Typocaps}

\LettrineTextFont{\itshape}
\usepackage{amsthm}

\newtheorem{definition}{Definition}[section]

\newtheorem{proposition}{Proposition}[section]
\newtheorem{remark}{Remark}[section]

\usepackage{amssymb}
\usepackage{pifont}
\newcommand{\xmark}{\ding{55}}%

\usepackage{minitoc}

\newcommand{\chapref}[1]{\hyperref[#1]{Chapter \ref{#1}}}
\newcommand{\secref}[1]{\hyperref[#1]{Section \ref{#1}}}

\usepackage{marginnote}

\usepackage[many]{tcolorbox}
\tcbuselibrary{breakable,xparse,skins}

\usepackage{newunicodechar}
\newunicodechar{λ}{{$\mathtt\lambda$}}
\newunicodechar{μ}{{$\mathtt\mu$}}

\usepackage{xparse}
\DeclareTColorBox{emphbox}{O{black}O{0cm}}{
    enhanced jigsaw,
    breakable,
    outer arc=0pt,
    arc=0pt,
    colback=white,
    rightrule=0pt,
    toprule=0pt,
    top=0pt,
    right=0pt,
    bottom=0pt,
    bottomrule=0pt,
    colframe=#1,
    enlarge left by=#2,
    width=\linewidth-#2,
}

\DeclareTColorBox{emphbox}{O{black}O{0cm}}{
    empty,
    breakable=true,
    outer arc=0pt,
    arc=0pt,
    rightrule=0pt,
    leftrule=2pt,
    borderline west={2pt}{0pt}{#1},
    toprule=0pt,
    top=0pt,
    right=-3pt,
    bottom=0pt,
    bottomrule=0pt,
    colframe=#1,
    enlarge left by=#2,
    width=\linewidth-#2,
}

\RequirePackage{algorithm}
\RequirePackage{algorithmic}
\usepackage{amsmath, amssymb, amsfonts, mathtools}
\usepackage{physics}
\usepackage{cancel}
\usepackage{thmtools, thm-restate}
\usepackage{accents}
\usepackage{bm}

\DeclareMathOperator{\diag}{diag}

\makeatletter
\newcommand{\ostar}{\mathbin{\mathpalette\make@circled *}}
\newcommand{\make@circled}[2]{%
  \ooalign{$\m@th#1\smallbigcirc{#1}$\cr\hidewidth$\m@th#1#2$\hidewidth\cr}%
}
\newcommand{\smallbigcirc}[1]{%
  \vcenter{\hbox{\scalebox{0.77778}{$\m@th#1\bigcirc$}}}%
}
\makeatother

\newcommand{\x}{\times}

\usepackage{pict2e, picture}
\makeatletter
\DeclareRobustCommand{\Arrow}[1][]{%
\check@mathfonts
\if\relax\detokenize{#1}\relax
\settowidth{\dimen@}{$\m@th\rightarrow$}%
\else
\setlength{\dimen@}{#1}%
\fi
\sbox\z@{\usefont{U}{lasy}{m}{n}\symbol{41}}%
\begin{picture}(\dimen@,\ht\z@)
\roundcap
\put(\dimexpr\dimen@-.7\wd\z@,0){\usebox\z@}
\put(0,\fontdimen22\textfont2){\line(1,0){\dimen@}}
\end{picture}%
}
\makeatother

\newcommand{\cA}{\mathcal{A}}

\newcommand{\cO}{\mathcal{O}}

\newcommand{\sA}{\mathsf{A}}
\newcommand{\sB}{\mathsf{B}}
\newcommand{\sC}{\mathsf{C}}
\newcommand{\sD}{\mathsf{D}}

\newcommand{\sH}{\mathsf{H}}

\newcommand{\sM}{\mathsf{M}}

\newcommand{\sS}{\mathsf{S}}

\newcommand{\sW}{\mathsf{W}}

\newcommand{\bC}{\mathbb{C}}

\newcommand{\R}{\mathbb{R}}

\newcommand{\bZ}{\mathbb{Z}}

\DeclareMathAlphabet{\nummathbb}{U}{BOONDOX-ds}{m}{n}

\makeatletter
\DeclareRobustCommand\widecheck[1]{{\mathpalette\@widecheck{#1}}}
\def\@widecheck#1#2{%
    \setbox\z@\hbox{\m@th$#1#2$}%
    \setbox\tw@\hbox{\m@th$#1%
       \widehat{%
          \vrule\@width\z@\@height\ht\z@
          \vrule\@height\z@\@width\wd\z@}$}%
    \dp\tw@-\ht\z@
    \@tempdima\ht\z@ \advance\@tempdima2\ht\tw@ \divide\@tempdima\thr@@
    \setbox\tw@\hbox{%
       \raise\@tempdima\hbox{\scalebox{1}[-1]{\lower\@tempdima\box
\tw@}}}%
    {\ooalign{\box\tw@ \cr \box\z@}}}
\makeatother

\begin{document}
\maketitle
 \begin{abstract}
    Recent advances in deep learning have relied heavily on the use of large Transformers due to their ability to learn at scale. However, the core building block of Transformers, the attention operator, exhibits quadratic cost in sequence length, limiting the amount of context accessible. Existing subquadratic methods based on low-rank and sparse approximations need to be combined with dense attention layers to match Transformers, indicating a gap in capability. In this work, we propose \textbf{Hyena}, a subquadratic drop-in replacement for attention constructed by interleaving implicitly parametrized \textbf{long convolutions} and \textbf{data-controlled gating}. In recall and reasoning tasks on sequences of thousands to hundreds of thousands of tokens, Hyena improves accuracy by more than $50$ points over operators relying on state-spaces and other implicit and explicit methods, matching attention-based models. We set a new state-of-the-art for dense-attention-free architectures on language modeling in standard datasets ({\sc WikiText103} and {\sc The Pile}), reaching Transformer quality with a $20\%$ reduction in training compute required at sequence length $2$K. Hyena operators are twice as fast as highly optimized attention at sequence length $8$K, and $100\x$ faster at sequence length $64$K.
\end{abstract}

\setlength\abovedisplayshortskip{2pt}
\setlength\belowdisplayshortskip{2pt}
\setlength\abovedisplayskip{2pt}
\setlength\belowdisplayskip{2pt}

\section{Introduction}
Large Transformers have enabled a number of breakthrough advances in modeling language, vision, audio, biology and numerous other domains \citep{vaswani2017attention}, \citep{dosovitskiy2020image}, \citep{radford2022robust}, \citep{cramer2021alphafold2}. Much of the success of Transformers, powered by the attention operator \citep{vaswani2017attention}, relies on their scaling properties \citep{hoffmann2022training} and the emergence of in-context learning \citep{garg2022can}, which allows them to generalize to unseen data and tasks given context as input. 
The Transformer block is a powerful tool for sequence modeling, but it is not without its limitations. One of the most notable is the computational cost, which grows rapidly as the length of the input sequence increases. Specifically, the cost scales quadratically with the length $L$ of the sequence, which places a strict limit on the amount of context that can be considered by the model.
Breaking the quadratic barrier is a key step towards new possibilities for deep learning, such as using entire textbooks as context, generating long-form music or processing gigapixel scale images.

Efforts to reduce the computational cost of attention in models primarily involve the use of linearized, low-rank, and sparse approximations \citep{child2019generating,wang2020linformer,kitaev2020reformer,zhai2021attention,roy2021efficient,schlag2021linear,tu2022maxvit}. These approaches introduce a trade-off between expressivity and speed, requiring hybridization with standard attention layers to reach Transformer quality \citep{mehta2022long,dao2022hungry}.

A growing amount of evidence suggests that attention mechanisms only utilize a small portion of their quadratic capabilities for language processing \citep{olsson2022context, dao2022hungry}, leading us to question its role as the gold-standard operator for deep learning at scale. Specifically, we ask:

\begin{figure*}[t]
    \centering
    \includegraphics[width=\linewidth]{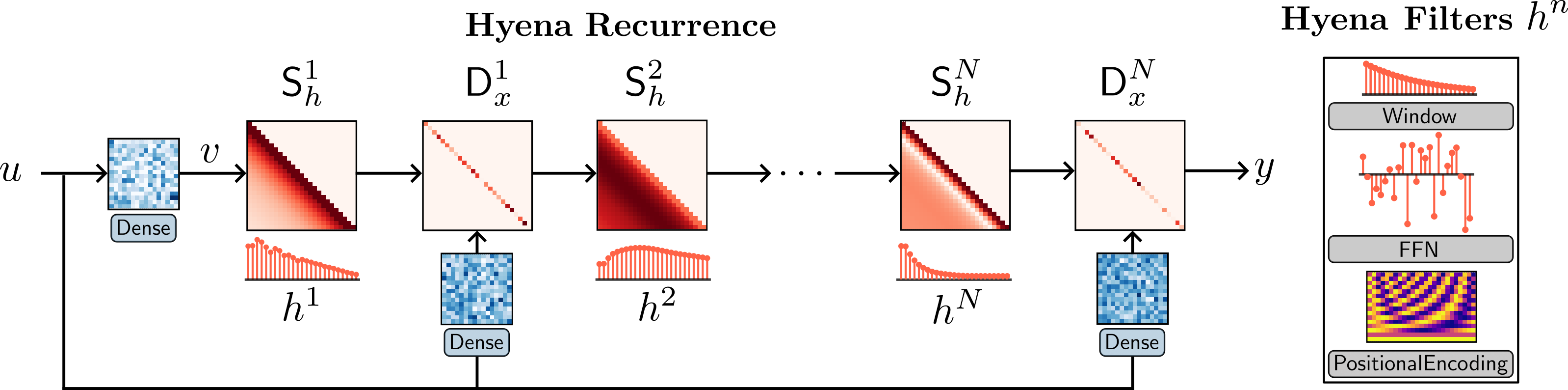}
    \vspace{-2mm}
    \caption{The ${\sf Hyena}$ operator is defined as a recurrence of two efficient subquadratic primitives: an implicit long convolution $h$ (i.e. {\sf Hyena} filters parameterized by a feed-forward network) and multiplicative element-wise gating of the (projected) input. The depth of the recurrence specifies the size of the operator. {\sf Hyena} can equivalently be expressed as a multiplication with \textit{data-controlled} (conditioned by the input $u$) diagonal matrices $\sD_x$ and Toeplitz matrices $\sS_h$. In addition, {\sf Hyena} exhibits sublinear parameter scaling (in sequence length) and unrestricted context, similar to attention, while having lower time complexity.}
    \label{arch}
\end{figure*}

{\centering
\textit{Are there subquadratic operators that can match the quality of attention at scale?}\par}

\vspace{0.5cm}

We obtain a positive answer based on a composition of efficient subquadratic primitives, such as \textit{element-wise multiplication} (gating) and \textit{long convolutions} i.e., convolutions with filter sizes as long as the input. We rely on a set of targeted reasoning tasks, grounded in recent work on \textit{mechanistic interpretability} \citep{elhage2021mathematical,power2022grokking,olsson2022context,zhang2022unveiling} such as recall and induction, to distill three properties of attention correlated with its performance and the quality gap with existing subquadratic approaches: 
\begin{itemize}[leftmargin=0.1in]
    \item[$a.$] \textbf{Data control:} Attention implements an expressive \textit{data-controlled} \citep{massaroli2020dissecting} linear operator\footnote{Self-attention can be expressed as $y = \sA(k, q) v$ where $\sA$ is the \textit{attention matrix} conditioned by linear projections $k, q$ of the input and multiplied by $v$, another projection.}, encoding an entire family of linear functions in a single block.
    \item[$b.$] \textbf{Sublinear parameter scaling:} Parameter counts of attention layers are decoupled from sequence length, allowing Transformers to allocate more parameters elsewhere e.g., the \textit{feed-forward neural networks} ({$\sf FFN$s}) between attention layers.
    \item[$c.$] \textbf{Unrestricted context:} For a given input, attention has an unrestricted context i.e., it can approximate dependencies between any two inputs, without arbitrary restrictions such as locality (except in cases using masking such as autoregressive models).
\end{itemize}
\paragraph{The ${\sf Hyena}$ hierarchy}
Guided by these findings, we introduce the ${\sf Hyena}$ hierarchy, an operator defined by a recurrence of two efficient subquadratic primitives: \textbf{a long convolution and element-wise multiplicative gating} (see Figure \ref{arch}). A specified depth (i.e., number of steps) of the recurrence controls the size of the operator. For short recurrences, existing models are recovered as special cases \citep{mehta2022long,dao2022hungry}. By mapping each step in the ${\sf Hyena}$ recurrence to its corresponding matrix form, we reveal ${\sf Hyena}$ operators to be equivalently defined as a decomposition of a \textit{data-controlled} matrix i.e., a matrix whose entries are functions of the input. Furthermore, we show how ${\sf Hyena}$ operators can be evaluated efficiently without materializing the full matrix, by leveraging fast convolution algorithms \citep{selesnick2017fast}. Empirically, ${\sf Hyena}$ operators are able to significantly shrink the quality gap with attention at scale, reaching similar perplexity and downstream performance with a smaller computational budget (Section \ref{res:lm}) and \textbf{without hybridization} of attention.

\paragraph{Narrowing the capabilities gap}
The design of {\sf Hyena} is motivated by a quality gap between standard dense attention and alternative subquadratic operators, which we identify by focusing on reasoning tasks correlated with language modeling performance at scale. We extend the suite of basic mechanistic interpretability benchmarks (\textit{induction} and \textit{recall}) with additional tasks that probe how quickly model performance degrades when task complexity increases (e.g. vocabulary size grows). In addition, we investigate the optimal parameterization of long convolutions in ${\sf Hyena}$. In the most challenging settings with hundreds of thousands of tokens, our implicit parameterization scheme improves over other operators leveraging state spaces \citep{gu2021efficiently}, frequency-domain parametrizations \citep{li2020fourier}, or standard convolutions by over $50\%$ accuracy.
\paragraph{Scaling in language and vision}
Next, we aim to verify whether rankings in our reasoning benchmark suite are predictive of quality at scale. We test ${\sf Hyena}$ on autoregressive language modeling at the sub-billion parameter scale, setting a new state-of-the-art for dense-attention-free architectures in standard datasets ({\sc WikiText103} and {\sc The Pile}) and matching Transformer quality. On the {\sc The Pile} at the $335$M parameter scale, we match Transformer perplexity with a $20\%$ reduction in the total count of \textit{floating point operations} (FLOPs). As an extension, we investigate the generality of ${\sf Hyena}$ operators by testing on large-scale image recognition, replacing attention in the Vision Transformer (ViT) \citep{dosovitskiy2020image}. In image classification, ${\sf Hyena}$ is able to match attention in accuracy when training on ImageNet-1k from scratch.
\paragraph{Toward much longer context}
Finally, we benchmark the efficiency of ${\sf Hyena}$ on long sequences. We measure $5$x speedups over dense self-attention at length $8192$ -- $2$x over highly optimized FlashAttention\footnote{FlashAttention is already 2-4x faster than a standard attention implementation in PyTorch.} \citep{dao2022flashattention} -- and $100$x speedup over FlashAttention at sequence lengths of $64$k, where standard attention implementation in PyTorch runs out of memory. 
\section{Preliminaries and Related Work}\label{back}
A discrete convolution is a function of two arguments: an input $u$ signal of length $L$ and a learnable filter $h$. The linear (aperiodic) convolution of a (possibly infinitely long) measurable\footnote{In the $L^1(\bZ)$ sense: $\sum_{t=-\infty}^\infty |h_t|<\infty$} filter $h$ with a length-$L$ input signal $u$ is defined as
\begin{equation}\label{eq:cnn}
    \begin{aligned} 
        y_t = (h * u)_t = \sum_{n=0}^{L-1} h_{t -n} u_n.
    \end{aligned}
\end{equation}

Generally, $u_t\in\R^D$ where $D$ is the width of the signal, or in deep learning parlance, the number of \textit{channels}. Without loss of generality, we specialize our analysis to \textit{single input single output} (SISO) layers, i.e. with $D=1$. The \textit{multiple input multiple output} (MIMO) case, canonical in standard convolutional layers, follows directly.

In this case, the input signal can be represented as a vector $u\in\R^L$ and the convolution as a matrix-vector product between the input and the Toeplitz kernel matrix $\sS_h \in \R^{L \times L}$ induced by the filter $h$:
\begin{equation}\label{eq:cnn_matvec}
    \begin{aligned} 
        (h * u) = 
        \begin{bmatrix}
            h_0 & h_{-1} & \cdots & h_{-L+1} \\
            h_1 & h_0 & \cdots & h_{-L+2} \\
            \vdots & \vdots & \ddots & \vdots \\
            h_{L-1} & h_{L-2} & \cdots & h_{0}
        \end{bmatrix}
        \begin{bmatrix}
            u_0\\
            u_1\\
            \vdots\\
            u_{L-1}
        \end{bmatrix}
    \end{aligned}
\end{equation}
\subsection{Explicit and Implicit Convolutions}
Parametrizing and optimizing convolution filters $h_t$ is a standard procedure in deep learning and more broadly signal processing. The classical approach of \textit{convolutional neural networks} (CNNs)  \citep{fukushima1982neocognitron,lecun1998gradient,ronneberger2015u,he2016deep} is to optimize directly the values $h_t$ of the filter's response at $M$ prescribed steps, a parametrization we call \textit{explicit}. $M$ is referred to as the \textit{filter size} and is typically much shorter than the input sequence length $M \ll L$. Such filters are denoted in signal processing as \textit{finite impulse response} (FIR). 

FIR filters are local and can capture dependencies between inputs separated at most by $M$ steps.
Their main advantage is their speed, with complexity $\mathcal{O}(ML)$. However, the number of parameters of FIR filters scales linearly with filter size, which can be computationally prohibitive. To disentangle the parameter count from the filter size, we can instead represent the filter $h_t$ as a parametric function of the time step $t$, i.e. $h_t = \gamma_\theta(t)$, where $\theta$ are the parameters of the function $\gamma_\theta$. This parametrization is called \textit{implicit}. The class of functions $\gamma_\theta$ is a design choice with a significant impact on the expressivity and computational complexity of the layer. 

One choice of implicit parametrization is to select $h$ as the response function of a linear state-space model (SSM) \citep{chen1984linear}, described by the first-order difference equation:
\begin{equation*}
    \begin{aligned}
        x_{t+1} &= \sA x_t + \sB u_t &&~~ \text{state equation} & \\
        y_t &= \sC x_t + \sD u_t &&~~ \text{output equation} &
    \end{aligned}
\end{equation*}
Here, the convenient choice of $x_0 = 0$ renders the input-output map to a simple convolution
\[
    \begin{aligned}
        y_t & =\sum_{n=0}^{t}\left(\sC\sA^{t - n}\sB + \sD \delta_{t-n}\right)u_n 
    \end{aligned}
\]
where $\delta_t$ denotes the Kronecker delta. We can then identify the filter $h$ as
\[
    t\mapsto h_t =
    \begin{cases}
        0 & t<0\\
        \sC \sA^t \sB + \sD\delta_t & t\geq 0
    \end{cases}
\]
where the entries of $\sA, \sB, \sC$ and $\sD$ are the learned parameters of the filter. In terms of layer design, the degrees of freedom of SSMs are the dimension of the state and the structure of the matrices. 
SSMs are a canonical example of how long convolutions with sub-linear parameter counts can improve deep learning models for long sequences \citep{gu2020hippo,gu2021efficiently}.
 Other implicit approaches include parametrizing filters as maps from (a positional encoding of) $t$ to the filter response i.e.  $\gamma_\theta : t \mapsto h_t=\gamma_\theta(t)$, for example with feed-forward neural networks \citep{romero2021ckconv,romero2021flexconv}. 
\begin{tcolorbox}[enhanced, drop fuzzy shadow, frame hidden, sharp corners, breakable, colback=blue!5] \textbf{Long convolutions and memory:} A crude proxy for \textit{memory} of a single computational unit is how far in the past it can access information to produce the output at a certain step. This can be roughly quantified by the number of non-zero entries $\partial y_t / \partial u_{t-n}$ for $n = 0, \ldots, t$. The memory of CNNs filters is equivalent to the filter size $M$ since $\partial y_t / \partial u_{t-n} = h_{n}$. The total mnemonic capacity of an all-convolutions CNN therefore scales with the number of model's parameters. Implicit parametrizations, on the other hand, allow us to disentangle the memory of each filter from the parameter count and where the length of the filter is implicitly controlled by the learned parameters. In an SSM, $\partial y_t / \partial u_{t-n} = \sC \sA^n \sB$ and the memory extent is solely determined by the spectral radius of $\sA$ and can be finely tuned by the training process\footnote{See e.g.\cite{gu2020hippo,gu2021efficiently}}. On the other hand, the number of parameters controls the \textit{expressivity} of the memory unit, e.g. the number of basis functions forming $h_t$. 
\end{tcolorbox}
\paragraph{Fast Methods for Convolutions}
One of the first applications of the Cooley-Tukey fast Fourier transform (FFT) algorithm was to implement convolution faster than the direct evaluation of \eqref{eq:cnn}.
At first glance \eqref{eq:cnn} comes with $O(L^2)$ an asymptotic time complexity. 
A common approach to achieve \textit{fast long convolutions} in subquadratic time is through the FFT algorithm. The method first converts the \textit{aperiodic} convolution into a \textit{circular} convolution \cite{selesnick2017fast} by appropriate zero-padding of input and filter sequences. The resulting kernel $\hat\sS_h$ is a circulant matrix and is diagonalized by the discrete Fourier basis
\[
    \hat \sS_h = \sW^{-1} \sD_{H} \sW
\]
where $\sW$ is the DFT matrix, $\sW_{tt'} = z^{-t}, z = e^{i2\pi t'/L}$ and $H$ is the DFT of the padded filter $h$, $H = \sW {\sf pad}(h)$. Thus, the calculation of such convolutions is performed as
\begin{equation*}
    \begin{aligned}
        {\sf pad}(y) &= \hat \sS_h {\sf pad}(u) \\
        &= \sW^{-1}\sD_H \sW ~{\sf pad}(u)\\
        &= {\sf iFFT}(\sD_H {\sf FFT}({\sf pad}(u)))
    \end{aligned}
\end{equation*}
where $\sD_H$ is the matrix with $\sW h$ on its diagonal. The above is known as the convolution theorem of DFT \citep{oppenheim1997signals}. In this ${\sf FFTConv}$ form the convolution can be performed \textbf{without materializing the operator $\sS_h$} with the same asymptotic cost $O(L\log_2 L)$ of FFT.
\subsection{The Self-Attention Operator}
At the heart of Transformers is the \textit{multi-head attention} (MHA) mechanism. Given a length-$L$ sequence $u\in\R^{L\times D}$, each \textit{head} of \textit{scaled self-attention} \citep{vaswani2017attention} is a map from $\R^{L\times D}$ to $\R^{L\times D}$ which performs the following operations
\begin{equation}\label{eq:att}
    \begin{aligned}
      \sA(u) &= {\sf SoftMax}\left(\tfrac{1}{\sqrt{D}}u \sM_q \sM^\top_k u^\top \right)\\
      y &= {\sf SelfAttention}(u) \\
        &= \sA(u)u \sM_v,
    \end{aligned}
\end{equation}
\begin{figure*}[t]
    \centering
    \includegraphics[width=0.69\linewidth]{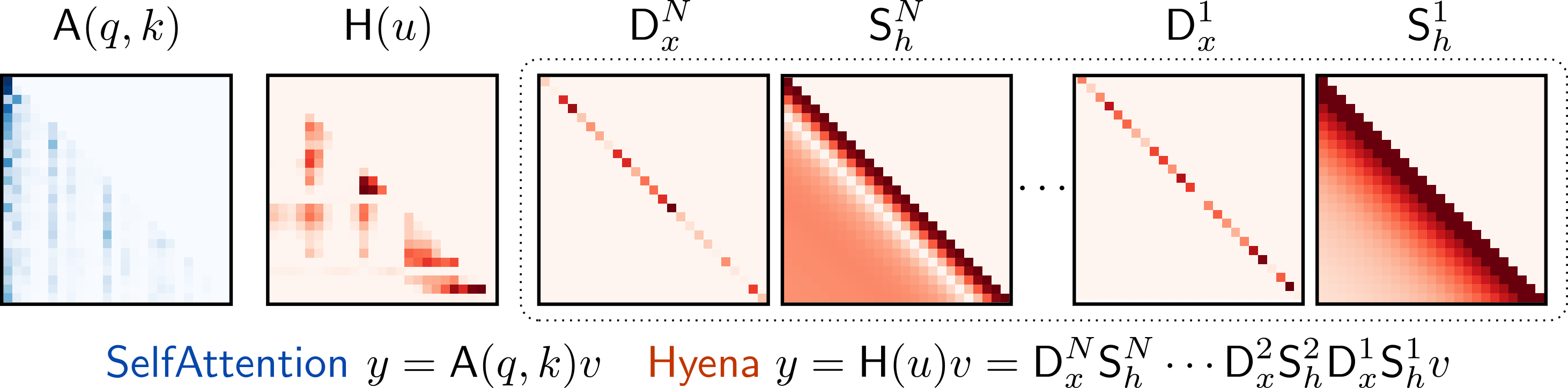}
    \vspace{-2mm}
    \caption{Comparison between data-controlled matrices: $\sf SelfAttention$ and $\sf Hyena$.}
    \label{fig:hyena_matrices}
\end{figure*}
where $\sM_q, \sM_k, \sM_v\in\R^{D\x D}$ are learnable linear projections and ${\sf SoftMax}$ is intended to be applied row-wise. Attention parametrizes a \textbf{family of dense linear operators} and for an input $u$, indexes through it via projections of $u$ i.e., $\sA(u)$. We refer to operators of this type as \textit{data-controlled}, as they encode a linear transformation $u \mapsto y$, that is, however, nonlinearly defined by $u$. This approach yields expressive nonlinear operators in $u$, and we hypothesize contributes, together with other mechanisms \citep{olsson2022context}, to the ability of certain operators to learn \textit{in-context} i.e., to adapt to unseen tasks by leveraging context. In deep learning, the projections take on specific names: \textit{query} $q=u\sM_q$, \textit{key} $k=u\sM_k$ and \textit{value} $v = u\sM_v$. We often rewrite the attention operator as $y = \sA(q,k)v$.
\begin{remark}
    Similarly to implicit convolutions, $\sf SelfAttention$ does not entangle its ability to access distant information with the number of parameters: it looks at the whole sequence at the price of $\cO(L^2)$ operations.
\end{remark}
\paragraph{Subquadratic Operators}
Existing approaches to subquadratic alternatives to attention can be summarized by altering the way the data control is implemented i.e., how the operator is nonlinearly defined by $u$, and then applied to $v$. For example, a layer of \textit{Attention-Free Transformers} (AFTs) \citep{zhai2021attention} constructs the operator through a combination of gating and {$\sf SoftMax$} (AFT full) or gating and a single explicit convolution (AFT conv). \textit{Gated State Spaces} (GSS) instead compose the operator via gating and a long convolution parametrized via SSMs. Taking this idea further, \textit{Hungry Hungry Hippo (H3)} \citep{dao2022hungry}, motivated by gaps of GSS on associative recall, extend the mechanism to include an additional gate and a short convolution obtained via a shift SSM. ${\sf Hyena}$ generalizes this body of work by introducing a recurrence of gates and implicit long convolutions, evaluated efficiently.
\section{Hyena: Definition and Properties}\label{allofhyena}
In this section, we define Hyena, a class of \textit{data-controlled} operators consisting of a recurrence of multiplicative gating interactions and long convolutions. Instead of seeking an approximation to attention, we guide our design by intentionally incorporating key computational properties of attention, including the decoupling of sequence length and parameter counts.
\subsection{${\sf Hyena}$ Recurrences}\label{hyena_op}
At a high level, ${\sf Hyena}$ consists of the following steps (setting $D = 1$ for clarity):
\begin{itemize}
    \item[$i.$] Compute a set of $N + 1$ linear projections of the input, similarly to attention. The number of projections $(v_t, x^1_t, \dots, x^N_t)$ need not be three. One projection takes the role of value, such that a linear input-output function can be defined as $y = \sH(u) v$ for some $\sH(u)$.
    \item[$ii.$] The matrix $\sH(u)$ is defined by interleaving implicit long convolutions and element-wise multiplication with one projection $x^i$ at a time, until all projections are exhausted. Evaluation of $\sH(u) v$ is done efficiently \textbf{without materializing} $\sH(u)$. By doing so, we implicitly define a data-controlled operator as a factorization of a matrix. The long convolutions forming $\sH(u)$ are parametrized implicitly to retain sublinear parameter scaling in sequence length.
\end{itemize}
Next, we formally define ${\sf Hyena}$, starting with its computational model. We leave the analysis of its data-controlled matrix form for the latter part of the section.
\begin{tcolorbox}[enhanced, sharp corners, drop fuzzy shadow, frame hidden, colback=yellow!15]
\begin{definition}[Order--$N$ $\sf Hyena$ Operator] 
Let $(v, x^1, \cdots, x^N)$ be projections of the input and let $h^1,\dots, h^N$ be a set of learnable filters. The ${\sf Hyena}_N$ operator is defined by the recurrence:
\begin{equation}\label{eq:hyena}
    \begin{aligned}
        z^1_t &={\ocra v}_t\\
        z^{n+1}_t & = x_t^n (h^n * z^n)_t & n=1,\dots,N\\
        {\color{blue!70}y}_t &= z_t^{N+1}
    \end{aligned} 
\end{equation}
\end{definition}
\end{tcolorbox}
\begin{remark}
The time complexity of a $\sf Hyena$ recurrence is $\mathcal{O}(N L \log_2 L)$. The input-output map can be rewritten as 
\[
y = x^N \cdot (h^N * ( x^{N-1} \cdot (h^{N-1} * (\cdots))))
\]
where each convolution is performed through the Fourier domain in $\mathcal{O}(L \log_2 L)$.
\end{remark}
Interestingly, the element-wise product in time domain corresponds to convolution in frequency domain, i.e.
\[ x_tu_t = (\hat x * \hat u)_t, \]
where $\hat x,\hat u$ denote the DFT of $x$ and $u$, respectively. Thus, $\sf Hyena$ is alternatively applying convolutions in the time and then the frequency domain (or alternatively applying element-wise products in the time and frequency domain). One potential explanation for the effectiveness of this procedure is that the convolution in the time domain (element-wise multiplication in the frequency domain) increases the memory length, allowing for a broader context to be taken into account. On the other hand, the element-wise multiplication in the time domain (convolution in the frequency domain) allows for more fine-grained selection of specific frequency components of the signal.
\subsection{${\sf Hyena}$ Matrices} 
${\sf Hyena}$ operators build on the ${\sf H3}$ mechanism developed by \citep{dao2022hungry}. For clarity of exposition, we once again consider the SISO case ($D=1$). Let $\sD_q$ and $\sD_k$ be the $L$-by-$L$ diagonal matrices whose respective main diagonal entries are the respective entries of $q$ and $k$. ${\sf H3}$ realizes a surrogate attention matrix with a data-controlled, parametrized decomposition in four terms:
\begin{equation}\label{eq:linear_attention}
    \begin{aligned}
        \sA(q,k) &= \sD_q \sS_\psi \sD_k \sS_\varphi \\
        {\sf H3}(q, k, v) &= \sA(q,k) v
    \end{aligned}
\end{equation}
where $\sS_\varphi,\sS_\psi$ are the Toeplitz matrices of learnable \textbf{causal} filters $\varphi,\psi$ parametrized via SSMs\footnote{For consistency with our discussion, we have swapped $k$ and $v$ compared to the notation in \citep{dao2022hungry}.}. Alongside the $qkv$-projections the filters constitute our degrees of freedom in the layer design. This decomposition allows evaluation of \eqref{eq:linear_attention} in just $\cO(L \log_2 L)$ time (two FFT convolutions and two element-wise products), i.e.
\begin{equation}
    \begin{aligned}
        z_{t} &= k_{t}(\varphi * v)_t \\
        y_{t} &= q_{t}(\psi * z)_t
    \end{aligned}
\end{equation}
$\small \sf Hyena$ represents a generalization of \eqref{eq:linear_attention} for an arbitrary number of projections -- not limited to three -- and with implicit free-form long filters for the convolutions. The resulting recurrence \eqref{eq:hyena} can be also represented in matrix form $y=\sH(u)v$. Let $\sD_x^n=\diag(x^n)\in\R^{L\x L}$ and let $\sS_h^n$ be the Toeplitz matrix corresponding to filter $h^n$. The resulting $\sf Hyena$ recurrence is linear in $v$ and can be rewritten in matrix form:
\[
    y = \sH(u)v = \sD_x^N\sS_h^N \cdots \sD_x^2\sS_h^2\sD_{x}^1\sS_h^1 v
\]
Figure~\ref{fig:hyena_matrices} visualizes an example decomposition.
\begin{remark}[$\sf Hyena$ generalizes $\sf H3$ and $\sf GSS$.]
    The $\sf H3$ mechanism \citep{dao2022hungry} corresponds to ${\sf Hyena}_2$ and $\sf GSS$ \citep{mehta2022long} is ${\sf Hyena}_{1}$, with a particular choice of parametrization for the long convolutions (SSMs).
\end{remark}
Analysis of the $\sf H3$ mechanism as a decomposition $\sD_q\sS_\psi\sD_k\sS_\varphi$ of its surrogate attention matrix\footnote{Some of this analysis is reported in the Appendix.} clarifies a connection to fast evaluation algorithms for matrix-vector multiplications. In particular, the generalization of \eqref{eq:linear_attention} to an arbitrary order is inspired by fast evaluation algorithms for structured dense matrices based on \textit{butterfly} decompositions \citep{li2015butterfly,dao2019learning, dao2022monarch}, with length of the decomposition closely tied to its expressivity (in the classes of matrices it can represent). The ${\sf Hyena}$ operator blends data control with a special case of butterfly decomposition.         
\begin{remark}
${\sf Hyena}$ operators have unbounded context. Namely, they are not artificially restricted by e.g., locality, and can learn long-range dependencies between any of the elements of $v$ via long convolutions, which we discuss next.
\end{remark}
\subsection{${\sf Hyena}$ Filters}\label{hyena_ker}
\begin{figure}[t]
    \centering
    \input{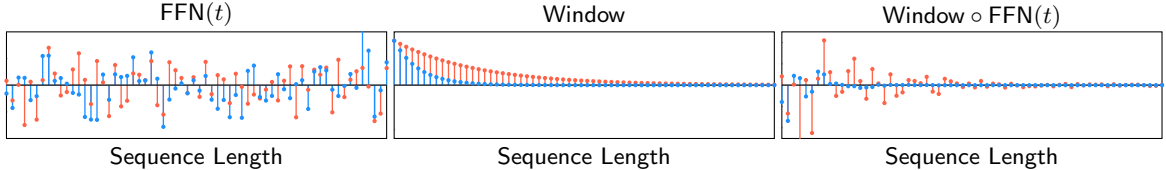}
    \vspace{-4mm}
    \caption{\textbf{[Top]:} Example of long convolution parametrization for ${\sf Hyena}$ operators, with a decay ${\sf Window}(t) = \exp\{- \alpha t\}$. Parameter $\alpha$ is modified across the independent channels of ${\sf Hyena}$ to regularize filters to be of different lengths. In practice, we add a bias term to our window, so that the filters are not constrained to be zeros after a length determined by the decay rate. 
    }
    \vspace{-4mm}
    \label{fig:modul}
\end{figure}

Here we provide details on the convolution parametrization. We represent the filters of each ${\sf Hyena}$ operator as a map from the time (or space) domain $t$ to values $h_t$, and learn it with a shallow feed-forward neural network ({$\sf FFN$}):
\begin{equation}\label{filt}
    h_t = {\sf Window}(t)\cdot({\sf FFN} \circ {\sf PositionalEncoding}) (t)
\end{equation}
This approach builds on the neural implicit representation literature \citep{mildenhall2021nerf,sitzmann2020implicit}, which has found application in long convolution layers \citep{romero2021ckconv, romero2021flexconv}. One advantage of \eqref{filt} is given by the decoupling of filter length and parameter cost. 
\paragraph{Specializing filters in Hyena}
The window and positional encoding functions are used to specialize filters in ${\sf Hyena}$ operators, biasing them towards a specific type. Figure \ref{fig:modul} provides an important example: we choose at least one of the convolutions in ${\sf Hyena}$ to be shaped towards exponential decay, mirroring the findings of \citep{li2022makes} in other applications.
Interestingly, we find that long exponentially decaying filters display synergy with high-frequency filters, as they enable the operator to select specific inputs at specific steps\footnote{This observation finds mirrors in the parametrization of the convolutions in H3 \citep{dao2022hungry} as a shift SSM and a diagonal SSM.}. Similarly to \citep{romero2021ckconv}, we use high-frequency periodic activations (sine) in the {$\sf FFN$}. This allows \eqref{filt} to learn filters with high-frequency content, addressing the low-frequency bias of neural networks \citep{basri2020frequency}.
Owing to the {$\sf FFN$}, the parametrization in \eqref{filt} can approximate filters obtained through other means, such as S4 \citep{gu2020hippo,gu2021efficiently}, CKConv \citep{romero2021ckconv}, SGConv \citep{li2022makes} and \textit{Fourier Neural Operator} (FNO) \citep{li2020fourier}.

\paragraph{Preserving causality}
Causality is necessary to train autoregressive language models, in order for the output at a given position to depend only on the past. For example, Transformers mask the attention matrix to be lower triangular. In the case of ${\sf Hyena}$, causality can be guaranteed by parametrizing causal convolutions: 
\begin{proposition}[Causal Hyenas]\label{prop:causality}
    If each filter $h^n, ~n=1,\dots, N$ is causal, then the corresponding ${\sf Hyena}_N$ operator is causal.
\end{proposition}
In practice, we need not constrain the learning of the filter \eqref{filt} to ensure its \textit{numerical} causality. If we use FFT-based convolution algorithms, all we need is to evaluate the filter at $t=0,\dots,L-1$ and zero-pad the input and filter sequences to $2 L - 1$ before taking FFT. 
\paragraph{Efficiency}
One bottleneck of long convolution models can be their low utilization of hardware accelerators, especially when they involve iterative numerical methods to materialize the filter\footnote{In contrast, deep learning primitives are designed for high GPU utilization, with {\sf FFNs} and attention usually reaching $50-70\%$ or higher, if optimized.}. Evaluation of \ref{filt} is fast, since it involves a single forward pass of an {$\sf FFN$}, and can be performed in parallel across sequence length and all orders of an ${\sf Hyena}$ operator as displayed in Algorithm \ref{alg:hyenaa}, increasing hardware utilization. An additional source of low utilization is the FFT, which is also shared by other long other convolutional layers. This bottleneck can be partially addressed by blocking \citep{selesnick2017fast}, and optimization of the underlying routines \citep{dao2022hungry}. We benchmark runtime in Section \ref{benchm}.

\subsection{{\sf Hyena} Algorithm}

A forward pass of {\sf Hyena} is summarized below. 

\setcounter{algorithm}{-1}
\begin{algorithm}[h]
    \caption{{\sf Projection}}\label{alg:hyenaa}
    \caption{{\sf Projection}}
    \begin{algorithmic}
    \REQUIRE Input sequence $u \in \R^{L \times D}$ \\ 
    \STATE {\small 1.} In parallel across $L$: $\hat{z} = {\sf Linear}(u)$, ${\sf Linear}:\R^{D} \rightarrow \R^{(N+1)D}$ \\
    \STATE {\small 2.} In parallel across $D$: $z = {\sf DepthwiseConv1d}(h, \hat{z} )$, $h$ is a short convolution filter \\
    \STATE {\small 3.} Reshape and split $z$ into $x^1, \dots, x^N, v$. Dimensions of one element are $x^n \in \R^{D \times L}$\\ 
    \STATE Return $x^1, \dots, x^N, v$, $x^n$
    \end{algorithmic}
\end{algorithm}

\setcounter{algorithm}{0}
\begin{algorithm}[h]
    \caption{${\sf HyenaFilter}$}\label{alg:hyenaa}
    \caption{${\sf Hyena}$ Filter}
    \begin{algorithmic}
    \REQUIRE Sequence length $L$, positional embedding dimension $D_e$ \\ 
    \STATE {\small 1.} $t  = {\sf PositionalEncoding}(L)$, $t \in \R^{L \times D_e}$ \\
    \STATE {\small 2.} In parallel across $N, L$: $\hat{h} = {\sf FFN}(t)$, ${\sf FFN}: \R^{D_e}\rightarrow \R^{N D}$, $\hat{h} \in \R^{L \times N D}$\\
    \STATE {\small 3.} Reshape to $\hat{h} \in \R^{N \times D \times L}$ \\
    \STATE {\small 4.} $h = \hat{h} \cdot {\sf Window}(t)$, $h \in \R^{N \times D \times L}$ \\
    \STATE {\small 5.} Split $h$ into $h^1, \dots, h^N$
    \STATE Return $h^1, \dots, h^N$
    \end{algorithmic}
\end{algorithm}

\setcounter{algorithm}{1}
\begin{algorithm}[h]
    \caption{${\sf Hyena}$ Operator}\label{alg:hyenaa}
    \caption{Forward pass of ${\sf Hyena}$}
    \begin{algorithmic}
    \REQUIRE Input sequence $u \in \R^{L \times D}$, order $N$, model width $D$, sequence length $L$, positional embedding dimension $D_e$ \\ 
    \STATE {\small 1.} $x^1, \dots, x^N, v = {\sf Projection}(u)$ \\
    \STATE {\small 2.} $h^1, \dots, h^N = {\sf HyenaFilter}(L, D_e)$ \\
    \FOR{$n = 1,\dots,N$}
    \STATE {\small 3.} In parallel across $D$: $v_{t} \leftarrow x^n_{t}\cdot {\sf FFTConv}(h^n, v)_{t}$ \\
    \ENDFOR  
    \STATE Return $y = v$
    \end{algorithmic}
\end{algorithm}

\begin{proposition}[Computational Complexity]
    The computational cost of processing an input $u\in\R^{L\x D}$ with an order-$N$ $\sf Hyena$ operator is
    \[
        \cO(N D L(\log_2 L + D))
    \]
\end{proposition}
\section{Experiments}\label{experiments}
\subsection{Shrinking the gap on in-context learning}\label{eval_ker}
We begin by empirically motivating the ${\sf Hyena}$ design, including the choice of long convolution parametrization. We consider the suite of tasks described in Table \ref{table:grok}.
\begin{table}[b]
    \small
    \centering
    \caption{A selection of our \textit{mechanistic design} benchmarks.}
    \setlength{\tabcolsep}{5.8pt}
    \label{table:grok}
    \begin{tabular}{@{}c|cc@{}}
    \toprule
    \textbf{Task} &\multicolumn{1}{c}{\textbf{Prompt}}&\multicolumn{1}{c}{\textbf{Target}}\\
    \midrule 
    Associative Recall & a, $1$, b, e, $3$, f, b & e\\  
    Majority & a, g, g, g, e, f, g & g \\
    Counting & a, b, b, b, a, c, b & 4\\
    ICL of Functions & $x_0$, $f(x_0)$, $\dots x_n$ & $f(x_n)$ \\ 
    Arithmetic & $1$, $3$, $5$, $+$, $6$, $8$, $3$ & $8$, $1$, $8$ \\
    \bottomrule
    \end{tabular}
\end{table}
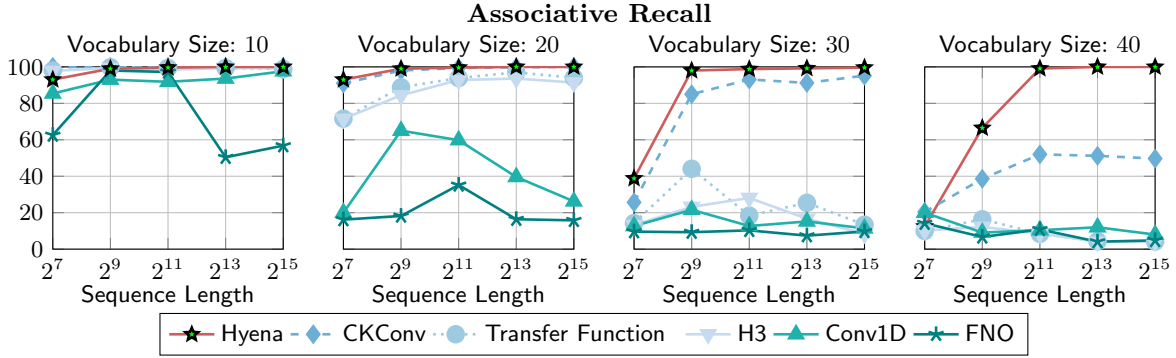
\begin{figure*}[t]
  \centering
  \begin{tikzpicture}[font=\small]

\definecolor{coral25314160}{RGB}{253,141,60}
\definecolor{cornflowerblue107174214}{RGB}{107,174,214}
\definecolor{darkgray176}{RGB}{176,176,176}
\definecolor{lightgray204}{RGB}{204,204,204}
\definecolor{lightsteelblue158202225}{RGB}{158,202,225}
\definecolor{orangered2308513}{RGB}{230,85,13}
\definecolor{powderblue198219239}{RGB}{198,219,239}
\definecolor{steelblue49130189}{RGB}{49,130,189}
\definecolor{brown}{RGB}{165,42,42}
\definecolor{cornflowerblue}{RGB}{100,149,237}
\definecolor{darkgray176}{RGB}{176,176,176}
\definecolor{indianred}{RGB}{205,92,92}
\definecolor{lightgray204}{RGB}{204,204,204}
\definecolor{teal}{RGB}{0,128,128}
\definecolor{lightseagreen}{RGB}{32,178,170}

\begin{groupplot}[group style={group size=4 by 1, horizontal sep=.8cm}]
\nextgroupplot[
width=.28\linewidth,
xlabel={$\sf Sequence Length$}, 
title={$\sf Vocabulary~Size$:~$10$}, title style={at={(.5,.92)}},
xlabel={$\sf Sequence~Length$}, xlabel style={at={(.5,-.15)}},
xmin=0, xmax=4,
ymin=0, ymax=100,
xtick={0,1,2,3,4}, xticklabels={$2^7$,$2^9$,$2^{11}$,$2^{13}$,$2^{15}$},
ytick={0,20,40,60,80,100}, yticklabels={0,20,40,60,80,100},
grid=both,
]
\addplot [line width=1pt, cornflowerblue107174214, dashed, mark=diamond*, mark size=3, mark options={solid}]
table {%
0 100
1 100
2 100
3 100
4 100
};
\addplot [line width=1pt, lightsteelblue158202225, dotted, mark=*, mark size=3, mark options={solid}]
table {%
0 98
1 99.8
2 99.4
3 99.4
4 98.8
};
\addplot [line width=1pt, powderblue198219239, mark=triangle*, mark size=3, mark options={solid,rotate=180}]
table {%
0 97.8
1 99.4
2 98.4
3 99.4
4 99.2
};
\addplot [line width=1pt, lightseagreen, mark=triangle*, mark size=3, mark options={solid}]
table {%
0 85.4
1 93
2 91.8
3 93.6
4 97.6
};
\addplot [line width=1pt, teal, mark=star, mark size=3, mark options={solid}]
table {%
0 62.6
1 98
2 97.2
3 50.4
4 56.8
};
\addplot [line width=1pt, indianred, mark=mystar, mark size=3, mark options={solid}]
table {%
0 93
1 99
2 99.6
3 100
4 100
};

\nextgroupplot[
width=.28\linewidth, 
title={$\sf Vocabulary~Size$:~$20$}, title style={at={(.5,.92)}},
xlabel={$\sf Sequence~Length$}, xlabel style={at={(.5,-.15)}},
xmin=0, xmax=4,
ymin=0, ymax=100,
xtick={0,1,2,3,4}, xticklabels={$2^7$,$2^9$,$2^{11}$,$2^{13}$,$2^{15}$},
ytick={0,20,40,60,80,100}, yticklabels={},
grid=both,
]

\addplot [line width=1pt, cornflowerblue107174214, dashed, mark=diamond*, mark size=3, mark options={solid}]
table {%
0 91
1 98
2 99.4
3 99.8
4 100
};
\addplot [line width=1pt, lightsteelblue158202225, dotted, mark=*, mark size=3, mark options={solid}]
table {%
0 71.6
1 88.8
2 93.9
3 97
4 94
};
\addplot [line width=1pt, powderblue198219239, mark=triangle*, mark size=3, mark options={solid,rotate=180}]
table {%
0 71.5
1 84.4
2 92.8
3 93.6
4 91.4
};
\addplot [line width=1pt, lightseagreen, mark=triangle*, mark size=3, mark options={solid}]
table {%
0 20
1 65
2 59.8
3 39.6
4 26.2
};
\addplot [line width=1pt, teal, mark=star, mark size=3, mark options={solid}]
table {%
0 16.2
1 18.2
2 35.2
3 16.4
4 15.8
};
\addplot [line width=1pt, indianred, mark=mystar, mark size=3, mark options={solid}]
table {%
0 93
1 99
2 99.6
3 100
4 100
};

\nextgroupplot[
width=.28\linewidth, 
title={$\sf Vocabulary~Size$:~$30$}, title style={at={(.5,.92)}},
xlabel={$\sf Sequence~Length$}, xlabel style={at={(.5,-.15)}},
xmin=0, xmax=4,
ymin=0, ymax=100,
xtick={0,1,2,3,4}, xticklabels={$2^7$,$2^9$,$2^{11}$,$2^{13}$,$2^{15}$},
ytick={0,20,40,60,80,100}, yticklabels={},
grid=both,
]
\addplot [line width=1pt, cornflowerblue107174214, dashed, mark=diamond*, mark size=3, mark options={solid}]
table {%
0 25.7
1 85
2 93.1
3 91.2
4 95.2
};
\addplot [line width=1pt, lightsteelblue158202225, dotted, mark=*, mark size=3, mark options={solid}]
table {%
0 14.3
1 44.1
2 18.4
3 25.4
4 13.4
};
\addplot [line width=1pt, powderblue198219239, mark=triangle*, mark size=3, mark options={solid,rotate=180}]
table {%
0 13.2
1 23.2
2 28.1
3 16.7
4 8.4
};
\addplot [line width=1pt, lightseagreen, mark=triangle*, mark size=3, mark options={solid}]
table {%
0 12.7
1 21.6
2 12.7
3 15.2
4 11.4
};
\addplot [line width=1pt, teal, mark=star, mark size=3, mark options={solid}]
table {%
0 9.6
1 9.3
2 10.3
3 7.4
4 9.7
};
\addplot [line width=1pt, indianred, mark=mystar, mark size=3, mark options={solid}]
table {%
0 38.8
1 98
2 98.8
3 99.2
4 99.6
};

\nextgroupplot[
width=.28\linewidth, 
title={$\sf Vocabulary~Size$:~$40$}, title style={at={(.5,.92)}},
xlabel={$\sf Sequence~Length$}, xlabel style={at={(.5,-.15)}},
xmin=0, xmax=4,
ymin=0, ymax=100,
xtick={0,1,2,3,4}, xticklabels={$2^7$,$2^9$,$2^{11}$,$2^{13}$,$2^{15}$},
ytick={0,20,40,60,80,100}, yticklabels={},
grid=both,
legend style = {
    draw=black,
    fill=none,
     /tikz/column 6/.style={column sep=5pt,},
    at={(0.45,-.35)},
},
legend columns=6
]
\addplot [line width=1pt, indianred, mark=mystar, mark size=3, mark options={solid}]
table {%
0 12.4
1 66.5
2 99.1
3 100
4 100
};\addlegendentry{$\sf Hyena$};

\addplot [line width=1pt, cornflowerblue107174214, dashed, mark=diamond*, mark size=3, mark options={solid}]
table {%
0 20.4
1 38.6
2 52
3 51.2
4 49.7
};\addlegendentry{$\sf CKConv$};
\addplot [line width=1pt, lightsteelblue158202225, dotted, mark=*, mark size=3, mark options={solid}]
table {%
0 10
1 16.4
2 8.4
3 4.3
4 4.2
};\addlegendentry{$\sf Transfer~Function$};
\addplot [line width=1pt, powderblue198219239, mark=triangle*, mark size=3, mark options={solid,rotate=180}]
table {%
0 10.2
1 12.2
2 8.6
3 4.1
4 4
};\addlegendentry{$\sf H3$};
\addplot [line width=1pt, lightseagreen, mark=triangle*, mark size=3, mark options={solid}]
table {%
0 20
1 9.1
2 10.4
3 12
4 7.99
};\addlegendentry{$\sf Conv1D$};
\addplot [line width=1pt, teal, mark=star, mark size=3, ]
table {%
0 14.2
1 6.7
2 10.8
3 4.1
4 4.8
};\addlegendentry{$\sf FNO$};
\end{groupplot}

\draw ({$(current bounding box.south west)!0.5!(current bounding box.south east)$}|-{$(current bounding box.south west)!1.1!(current bounding box.north west)$}) node[
  scale=1.1,
  anchor=north,
  text=black,
  rotate=0.0
]{\textbf{Associative~Recall}};
\end{tikzpicture}
  \caption{Benchmark of long convolution parametrizations in order $2$ ${\sf Hyena}$ operators on associative recall (\%). Our results show that implicit parametrizations scale more favorably in vocabulary size (number of possible values of tokens in the input) and length of the sequence.}
  \label{fig:synthetics1}
\end{figure*}
Our evaluation is grounded in recent work on mechanistic interpretability of Transformers \citep{elhage2021mathematical,power2022grokking,olsson2022context,zhang2022unveiling}. Recently, associative recall, in particular, has been successfully used to guide the design of H3 \citep{dao2022hungry}. We extend the suite of tasks from these works and include benchmarking more challenging versions of each task . For example, solving associative recall with a vocabulary size of only $10$ reveals whether a model is structurally capable of performing recall. Testing on much longer sequences and larger vocabularies reveals additional gaps in performance that are otherwise hidden.
\paragraph{How to parametrize long convolutions}
We compare the performance of the following long convolution parametrizations for $S^1$ and $S^2$ in an order $2$ Hyena:
\begin{itemize}[leftmargin=0.1in]
    \item Conv1d: Explicit convolutions (regular convolution layers with fixed filter size).
    \item FNO: Filters parametrized explicitly in the frequency-domain \citep{li2020fourier}.
    \item H3: Implicit parametrization using state-space models (SSMs), in particular the standard S4 \citep{gu2021efficiently}. 
    \item TransferFunc: Implicit parametrization via transfer functions, a classical system-theoretic generalization of SSMs\footnote{Transfer functions roughly correspond to a frequency-domain representation of SSMs.} 
    \item CKConv: Implicit parametrization using {$\sf FFN$s} \citep{romero2021ckconv}. 
    \item ${\sf Hyena}$: Combination of implicit parametrizations via {$\sf FFN$s} (with exponential decay modulation as shown in Figure \ref{fig:modul}), and short explicit filters.
\end{itemize}
All models have the same width and $2$ layers. Figure \ref{fig:synthetics1} shows implicit approaches based on {\sf FFNs} outperform other long convolutions, with the gap widening on longer sequences and larger vocabulary sizes. We train a different model on each setting of sequence length and vocabulary size. The ranking is correlated with the ability to decouple sequence length from parameter count (${\sf Hyena}$, CKConv, TransferFunc, H3) and expressivity (Hyena, CKConv). We observe similar trends on the other tasks.
\paragraph{Pushing sequence length to the limit}
Next, we evaluate associative recall performance on extremely long sequences of length $131$k. To the best of our knowledge, these represent the first empirical display of attention-free in-context learning on sequences of this length. The gap between parametrization schemes widens as shown in Appendix A, with ${\sf Hyena}$ outperforming CKConv by $80$ points.
\paragraph{Comparing operators}
We repeat our associative recall experiment, this time benchmarking different $2$ layer models rather than changing the convolution parametrization: an order $2$ Hyena, GSS \citep{mehta2022long}, H3 \citep{dao2022hungry}, AFT-conv \citep{zhai2021attention}, RWKV \citep{PENG_RWKV-LM_2021}, and a standard GPT \citep{brown2020language} using FlashAttention \citep{dao2022flashattention}. As shown in Table \ref{table:synthetic2}, ${\sf Hyena}$ is the only operator able to solve the task. Our results challenge the observation that only Transformers are capable of challenging in-context learning.
\begin{table}[t]
\small
\centering
\caption{Test accuracy (\%) for associative recall on longer sequences, vocabulary size $30$. The symbol \xmark~is used to mark settings where the model does not fit in memory.}
\label{table:synthetic2}
\setlength{\tabcolsep}{4pt}
\begin{tabular}{@{}c|c|c|c|ccccccc@{}}
\toprule
Sequence length & ${\sf Hyena}$ & FlashTransformer & Transformer &\multicolumn{1}{c}{GSS}  &\multicolumn{1}{c}{H3} & \multicolumn{1}{c} {AFT} & \multicolumn{1}{c}{RWKV} \\
\midrule 
$30$k & $100.0$ & $32.4$ & \xmark & $5.3$ & $8.4$ & $2.3$ & $12.4$ \\
$64$k & $100.0$ & $26.7$ & \xmark & $2.1$ & $4.3$ & $1.2$ & $6.5$ \\
$131$k & $97.2$ & \xmark & \xmark & $0.1$ & $0.6$ & $0.8$ & $2.3$ \\
\bottomrule
\end{tabular}
\end{table}
Surprisingly, rankings of model performance at a fixed sequence length on {\sf The Pile} are consistent with rankings on aggregate scores on our synthetics (Appendix \ref{app:add_results}).
\paragraph{Generality of ${\sf Hyena}$ operators and filters}
${\sf Hyena}$ operators and filters can also applied successfully beyond language tasks. We experiment on sequential CIFAR, where pixels are flattened as a sequence, and use the same operator defined for language. We reach the accuracy of standard S4 \citep{gu2021efficiently} with same model size ($91\%$). In Section \ref{sec:image_classification} and Appendix \ref{appendix:experiment-details}, we discuss larger-scale image classification experiments with Hyena.
\subsection{Language Modeling}\label{res:lm}
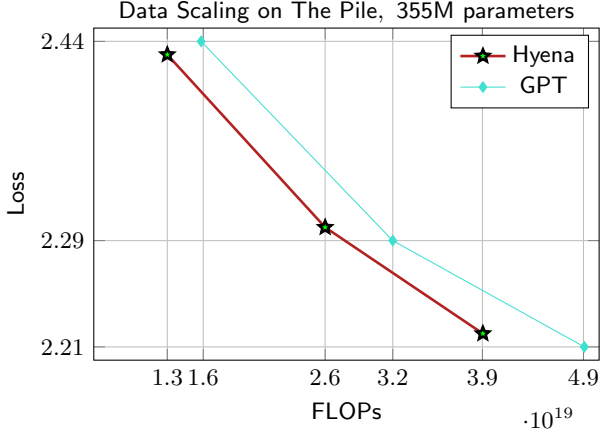
\begin{figure}[t]
    \centering
\begin{tikzpicture}[font=\small]

\definecolor{brown}{RGB}{165,42,42}
\definecolor{darkgray176}{RGB}{176,176,176}
\definecolor{firebrick}{RGB}{178,34,34}
\definecolor{indianred}{RGB}{205,92,92}
\definecolor{lightgray204}{RGB}{204,204,204}
\definecolor{lightseagreen}{RGB}{32,178,170}
\definecolor{teal}{RGB}{0,128,128}
\definecolor{turquoise}{RGB}{64,224,208}

\begin{axis}[
width=8.3cm, height=6cm,
title={$\small{{\sf Data~Scaling~on~The~Pile,~355M~parameters}}$},
title style = {at={(.5, .94)}},
xlabel={$\small \sf FLOPs$}, ylabel={$\small\sf Loss$}, 
ylabel style={at={(-.12,.5)}},
xmin=7e+18, xmax=4.9e+19, ymin=2.2, ymax=2.45,
major grid style={lightgray,thin},
minor grid style={lightgray,very thin},
grid=both,
xtick={1.311e+19,1.61e+19,2.623e+19,3.185e+19,3.93e+19,4.77e+19},
xticklabels={
  \(\displaystyle {1.3}\),
  \(\displaystyle {1.6}\),
  \(\displaystyle {2.6}\),
  \(\displaystyle {3.2}\),
  \(\displaystyle {3.9}\),
  \(\displaystyle {4.9}\)
},
ytick={2.44, 2.29, 2.21},
yticklabels={
  \(\displaystyle {2.44}\),
  \(\displaystyle {2.29}\),
  \(\displaystyle {2.21}\)
}
]
\addplot [mark=mystar, line width=1pt, firebrick]
table{%
x  y
1.311e+19 2.43
2.623e+19 2.30
3.934e+19 2.22
};\addlegendentry{$\sf Hyena$};
\addplot [draw=black, mark=diamond*, turquoise]
table{%
x  y
1.592e+19 2.44
3.185e+19 2.29
4.777e+19 2.21
};\addlegendentry{$\sf GPT$};
\end{axis}

\end{tikzpicture}
    \vspace{-4mm}
    \caption{Preliminary "scaling law" of language models on {\sc The Pile}. Comparison of our approach (red) based on long convolutions and gating (${\sf Hyena}$) and a standard GPT (blue) \citep{brown2020language}. We reach perplexity of GPT with a smaller training FLOP budget.}
    \label{fig:hyena_laws}
\end{figure}

Next, we verify the scaling of ${\sf Hyena}$ on autoregressive language modeling. We evaluate the perplexity on {\sc WikiText103} (Table \ref{wt103}) and {\sc The Pile} (Table \ref{pile}). On the {\sc The Pile}, we train different models for $5, 10, 15$ billion tokens (different runs), adjusting the learning rate scheduler. ${\sf Hyena}$ is the first attention-free, convolution architecture to match GPT quality with a $~20\%$\footnote{The FLOP reduction consists in the \textit{non-parametric} FLOPs of {$\sf SelfAttention$} devoted to attention matrix computation. The ratio of parametric to non-parametric FLOPs (and hence the gains) depend on the ratio of model width $D$ and sequence length $L$ used in training.} reduction in total FLOPs. Preliminary scaling laws are shown in Figure \ref{fig:hyena_laws}, collecting the training runs at $5, 10, 15$ billion tokens. Each curve represents a different training run.
In Appendix \ref{appendix:experiment-details}, we provide results on the PG-19 long-range benchmark \citep{raecompressive2019}.

\begin{table}[t]
    \small
    \centering
    \begin{minipage}{.38\textwidth}
        \centering
        \caption{Perplexity on {\sc WikiText103} (same tokenizer). $^*$ are results from \citep{dao2022hungry}. Deeper and thinner models (Hyena-slim) achieve lower perplexity.}
        \vspace{2mm}
        \label{wt103}
        \setlength{\tabcolsep}{4pt}
        \begin{tabular}{@{}c|cc@{}}
        \toprule
        Model &\multicolumn{1}{c}{{\sc Perplexity}} \\
        \midrule 
        Transformer ($125$M) & $18.6$ & \\
        Hybrid H3 ($125$M) & $18.5^*$ \\
        Performer ($125$M) & $26.8^*$  \\ 
        Reformer  ($125$M)& $25.6^*$  \\ 
        \midrule
        AFT-conv ($125$M) & $28.2$ \\
        Linear Attention ($125$M) & $25.6^*$ \\
        \midrule
        ${\sf Hyena}$-$3$ ($125$M) & $18.6$ \\
        ${\sf Hyena}$-$3$-slim ($125$M) & $18.5$ \\
        \bottomrule
        \end{tabular}
    \end{minipage}
    \hspace{0.7cm}
    \begin{minipage}{.55\textwidth}
        \centering
        \vspace{-18mm}
        \caption{Perplexity on {\sc The Pile} for models trained until a total number of tokens e.g., $5$ billion (different runs for each token total). All models use the same tokenizer (GPT2). FLOP count is for the $15$ billion token run.}
        \vspace{2mm}
        \label{pile}
        \setlength{\tabcolsep}{4pt}
        \begin{tabular}{@{}c|cccc@{}}
        \toprule
        Model &\multicolumn{1}{c}{{\sc $5$B}} & \multicolumn{1}{c}{{\sc $10$B}} & \multicolumn{1}{c}{{\sc $15$B}} & \multicolumn{1}{c}{{\sc FLOPs ($10^{19}$)}}\\
        \midrule 
        GPT ($125$M) & $13.3$ & $11.9$ & $11.2$ & $1.88$ \\
        ${\sf Hyena}$-$2$ ($153$M)& $13.3$ & $11.8$ & $11.1$ & $\textbf{1.87}$ \\
        \midrule
        GPT ($355$M) & $11.4$ & $9.8$ & $9.1$ & $4.77$ \\
        ${\sf Hyena}$-$2$ ($355$M) & $11.3$ & $9.8$ & $9.2$ & $\textbf{3.93}$ \\
        \bottomrule
        \end{tabular}
    \end{minipage}
\end{table}


\subsection{Downstream Evaluation}
We perform a downstream evaluation on SuperGLUE \citep{wang2019superglue} tasks. We compare ${\sf Hyena}$ (trained for $137$ billion tokens) with the best available pre-trained attention-free model, RWKV \citep{PENG_RWKV-LM_2021} (trained for $332$ billion tokens), and a reference GPTNeo \citep{gpt-neo} (trained for $300$ billion tokens) of the same size. Tables \ref{supergluezero} and \ref{supergluefew} summarize the results. ${\sf Hyena}$ performs similarly to other models despite having been trained on less than half the number of total tokens. We observe ${\sf Hyena}$ to display characteristic few-shot capabilities of standard Transformers, with some tasks e.g., MultiRC seeing a lift of more than $20\%$ accuracy over zero-shot when the model is provided additional prompts as context. The improvements are more noticeable in generation tasks, where the additional prompts can instruct the model on how it should be responding to the questions. We report an additional downstream evaluation on the LAMBADA  task \citep{paperno2016lambada} in Appendix \ref{appendix:experiment-details}.
\begin{table}[!h]
\small
\centering
\caption{Zero-shot accuracy ($\%$) on {\sc SuperGLUE} tasks for small models.}
\label{supergluezero}
\vspace{2mm}
\setlength{\tabcolsep}{4pt}
\begin{tabular}{@{}c|ccccccccc@{}}
\toprule
Model &\multicolumn{1}{c}{{\sc WSC}} & \multicolumn{1}{c}{{\sc WIC}} & \multicolumn{1}{c}{{\sc RTE}} & \multicolumn{1}{c}{{\sc CB}} & \multicolumn{1}{c}{{\sc MultiRC}} &  \multicolumn{1}{c}{{\sc ReCoRD}} &  \multicolumn{1}{c}{{\sc BoolQ}} &  \multicolumn{1}{c}{{\sc COPA}} & \multicolumn{1}{c}{{\sc Average}} \\
\midrule 
GPTNeo \citep{gpt-neo} & $\mathbf{27.9}$ & $50.0$ & $45.1$ & $\mathbf{41.1}$ & $0.0$ & $\mathbf{61.7}$ & $\mathbf{62.2}$ & $62.0$ & $\mathbf{43.8}$ \\ 
RWKV \citep{PENG_RWKV-LM_2021} & $13.4$ & $\mathbf{52.3}$ & $\mathbf{46.9}$ & $25.0$ & $0.0$ & $58.5$ & $\underline{59.2}$ & $\underline{66.0}$ & $40.2$ \\
${\sf Hyena}$ & $\underline{21.2}$ & $\underline{50.5}$ & $\underline{46.6}$ & $\underline{39.3}$ & $\mathbf{1.1}$ & $\underline{59.4}$ & $51.8$ & $\mathbf{70.0}$ & $\underline{41.5}$ \\
\bottomrule
\end{tabular}
\end{table}
\begin{table}[!h]
\small
\centering
\caption{Few-shot ($3$) accuracy ($\%$) on {\sc SuperGLUE} tasks for small models.}
\label{supergluefew}
\vspace{2mm}
\setlength{\tabcolsep}{4pt}
\begin{tabular}{@{}c|ccccccccc@{}}
\toprule
Model &\multicolumn{1}{c}{{\sc WSC}} & \multicolumn{1}{c}{{\sc WIC}} & \multicolumn{1}{c}{{\sc RTE}} & \multicolumn{1}{c}{{\sc CB}} & \multicolumn{1}{c}{{\sc MultiRC}} &  \multicolumn{1}{c}{{\sc ReCoRD}} &  \multicolumn{1}{c}{{\sc BoolQ}} &  \multicolumn{1}{c}{{\sc COPA}} & \multicolumn{1}{c}{{\sc Average}} \\
\midrule 
GPTNeo \citep{gpt-neo} & $\underline{38.5}$ & $\underline{50.0}$ & $\mathbf{53.8}$ & $\underline{42.9}$ & $\underline{22.4}$ & $\mathbf{61.4}$ & $\mathbf{61.0}$ & $63.0$ & $\underline{49.1}$ \\ 
RWKV \citep{PENG_RWKV-LM_2021} & $32.7$ & $49.4$ & $47.2$ & $37.5$ & $0.0$ & $\underline{58.3}$ & $55.0$ & $\underline{64.0}$ & $43.0$ \\
${\sf Hyena}$ & $\mathbf{39.4}$ & $\mathbf{50.1}$ & $\underline{47.6}$ & $\mathbf{46.4}$ & $\mathbf{26.7}$ & $58.1$ & $\underline{56.0}$ & $\mathbf{70.0}$ & $\mathbf{49.3}$ \\
\bottomrule
\end{tabular}
\end{table}
\subsection{Benchmarking}
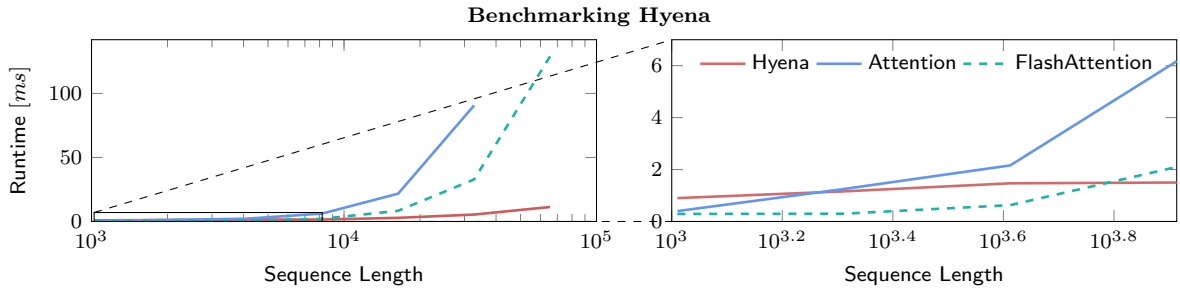
\begin{figure}[t]
    \centering
    \begin{tikzpicture}[font=\footnotesize
]
\definecolor{brown}{RGB}{165,42,42}
\definecolor{cornflowerblue}{RGB}{100,149,237}
\definecolor{darkgray176}{RGB}{176,176,176}
\definecolor{indianred}{RGB}{205,92,92}
\definecolor{lightgray204}{RGB}{204,204,204}
\definecolor{teal}{RGB}{0,128,128}
\definecolor{lightseagreen}{RGB}{32,178,170}
 \begin{axis}[
   width=.5\linewidth, height=4cm,
   xmode=log,
   xmin=1000, xmax = 100000,
   ymin=0,
   xlabel=In,
   ylabel=out,
   name=ax1,
   xlabel={$\sf Sequence~Length$},
   ylabel={${\sf Runtime}~[ms]$},
 ]
\addplot [line width=1pt, indianred]
table {%
1024 0.9
2048 1.16
4096 1.47
8192 1.5
16384 2.84
32768 5.41
65536 11.32
};
\addplot [line width=1pt, cornflowerblue]
table {%
1024 0.4
2048 1.25
4096 2.16
8192 6.17
16384 21.74
32768 90.71
};
\addplot [line width=1pt, lightseagreen, dashed]
table {%
1024 0.29
2048 0.3
4096 0.63
8192 2.1
16384 8.33
32768 32.85
65536 129.07
};

  \coordinate (c1) at (axis cs:1024,0);
  \coordinate (c2) at (axis cs:1024,7);
  \draw (c1) rectangle (axis cs:8196,7);

\end{axis}
\begin{axis}[
    width=.5\linewidth, height=4cm,
    name=ax2,
    xmin=1000,xmax=8192,
    ymin=0,ymax=7,
    xlabel={${\sf Sequence~Length}$},
    xmode=log,
    xminorgrids=true,
    legend style={font=\footnotesize, draw=none, fill=none, align=left},
    legend columns = 3,
    legend cell align={left},
    at={($(ax1.south east)+(1cm,0)$)},
    clip=false
 ]
    \addplot [line width=1pt, indianred, opacity=.9]
        table {%
        1024 0.9
        2048 1.16
        4096 1.47
        8192 1.5
    };\addlegendentry{$\sf Hyena$}
    \addplot [line width=1pt, cornflowerblue]
        table {%
        1024 0.4
        2048 1.25
        4096 2.16
        8192 6.17
    };\addlegendentry{$\sf Attention$}
    \addplot [line width=1pt, lightseagreen, dashed]
        table {%
        1024 0.29
        2048 0.3
        4096 0.63
        8192 2.1
    };\addlegendentry{$\sf FlashAttention$}
\end{axis}

\draw [dashed] (c1) -- (ax2.south west);
\draw [dashed] (c2) -- (ax2.north west);

\draw ({$(current bounding box.south west)!0.5!(current bounding box.south east)$}|-{$(current bounding box.south west)!1.16!(current bounding box.north west)$}) node[
  scale=1,
  anchor=north,
  text=black,
  rotate=0.0
]{\textbf{Benchmarking~Hyena}};
\end{tikzpicture}
    \vspace{-5mm}
    \caption{Benchmarking runtime of Hyena, Attention and FlashAttention with varying sequence lengths. Batch size is set to $64$. The figure on the right is an inset showing a zoomed-in portion of the figure on the left.}
    \label{fig:benchmarking_plot}
\end{figure}

We benchmark runtime of an order $2$ ${\sf Hyena}$ operator compared to attention and FlashAttention layers \citep{dao2022flashattention}. ${\sf Hyena}$ uses a fused CUDA kernel to perform ${\sf FFTConv}$ \citep{dao2022hungry}. We set batch size to $64$ and measure runtime (in milliseconds). Results are provided in Figure \ref{fig:benchmarking_plot}. ${\sf Hyena}$ speedups reach $100\x$ at sequence length $64$K. Crossover points for ${\sf Hyena}$ and attention is at length $2048$, and for ${\sf Hyena}$ and FlashAttention is between $4096$ and $8196$. Despite the absolute reduction in FLOPs, speedups are achieved only on longer sequences when the gap grows sufficiently large. This occurs because hardware utilization of ${\sf Hyena}$ is lower than FlashAttention. We expect the gap between theoretical maximum speedup to shrink with improved implementations of ${\sf FFTConv}$ and specialized hardware.

\subsection{Large-Scale Image Classification}\label{benchm}
\label{sec:image_classification}
Finally, we demonstrate the potential of ${\sf Hyena}$ as a general deep learning operator by applying it to image classification. On {\sf ImageNet}, we drop-in replace attention layers in the \textit{Vision Transformer} (ViT) \citep{dosovitskiy2020image} with the ${\sf Hyena}$ operator (without changes from its language counterpart) and match performance with ViT. We also show that using smaller image patches boosts performance in both attention and ${\sf Hyena}$. Since this results in longer sequence lengths, we expect ${\sf Hyena}$ to outperform in speed as patches get more fine-grained approaching pixel-level. On CIFAR-2D, we test a 2D version of ${\sf Hyena}$ long convolution filters in a standard convolutional architecture, which improves on the 2D long convolutional model S4ND \citep{nguyen2022s4nd} in accuracy with a $8\%$ speedup and 25\% fewer parameters. See Appendix \ref{appendix:image-classification} for additional vision architectures and training procedure details.


\begin{table}[h]
\small
\centering
\caption{Image classification top-1 accuracy.}
\label{image_results}
\setlength{\tabcolsep}{4pt}
\begin{tabular}{@{}c|cccc@{}}
\toprule
Model & \multicolumn{1}{c} {{\sc Patch Size}} &  \multicolumn{1}{c} {{\sc Seq Len}} & \multicolumn{1}{c} {{\sc Dataset}} & \multicolumn{1}{c}{{\sc Acc (\%)}} \\
\midrule 
ViT ($87$M) & 16x16 & 196 & ImageNet-1k & 78.5 \\
Hyena-ViT ($88$M) & 16x16 & 196 & ImageNet-1k & 78.5 \\
\midrule
ViT ($87$M) & 8x8 & 1024 & ImageNet-1k & 80.0 \\
Hyena-ViT ($88$M) & 8x8 & 1024 & ImageNet-1k & 79.8 \\
\midrule
S4ND-ISO ($268$k) & - & - & CIFAR-10 & 89.9 \\
Hyena-ISO ($202$k) & - & - & CIFAR-10 & 91.2 \\
\bottomrule
\end{tabular}
\end{table}
\section{Discussion and Conclusion}
In this work, we introduced an attention-free drop-in replacement to the core building block of many large-scale language models. ${\sf Hyena}$ operators are a recurrence of gating and implicitly parametrized long convolutions, can be evaluated efficiently in subquadratic time, and can learn in-context on very long sequences. 
On {\sc The Pile}, deep stacks of ${\sf Hyena}$ operators constitute one of the first attention-free, convolutional architectures to match perplexity and downstream performance of Transformers with a significant reduction in training compute. 
Our promising results at the sub-billion parameter scale suggest that attention may not be all we need, and that simpler subquadratic designs such as ${\sf Hyena}$, informed by a set of simple guiding principles and evaluation on mechanistic interpretability benchmarks, may form the basis for efficient large models. We are excited about what new capabilities {\sf Hyena} opens up as we scale and optimize the inference speed of these models.
\section*{Acknowledgments}

We would like to thank Karan Goel, Albert Gu, Avanika Narayan, Khaled Saab, Michael Zhang, Elliot Epstein and Sabri Eyuboglu for helpful discussion and feedback on earlier drafts, and Together Computer and Crusoe for providing the compute used to train models in this paper. We gratefully acknowledge the support of NIH under No. U54EB020405 (Mobilize), NSF under Nos. CCF1763315 (Beyond Sparsity), CCF1563078 (Volume to Velocity), and 1937301 (RTML); US DEVCOM ARL under No. W911NF-21-2-0251 (Interactive Human-AI Teaming); ONR under No. N000141712266 (Unifying Weak Supervision); ONR N00014-20-1-2480: Understanding and Applying Non-Euclidean Geometry in Machine Learning; N000142012275 (NEPTUNE); NXP, Xilinx, LETI-CEA, Intel, IBM, Microsoft, NEC, Toshiba, TSMC, ARM, Hitachi, BASF, Accenture, Ericsson, Qualcomm, Analog Devices, Google Cloud, Salesforce, Total, the HAI-GCP Cloud Credits for Research program,  the Stanford Data Science Initiative (SDSI), Department of Defense (DoD) through the National Defense Science and Engineering Graduate Fellowship (NDSEG) Program, and members of the Stanford DAWN project: Facebook, Google, and VMWare. This work is supported by NSF (1651565), AFOSR (FA95501910024), ARO (W911NF-21-1-0125), ONR, DOE (DE-SC0022222), CZ Biohub, and Sloan Fellowship. The U.S. Government is authorized to reproduce and distribute reprints for Governmental purposes notwithstanding any copyright notation thereon. Any opinions, findings, and conclusions or recommendations expressed in this material are those of the authors and do not necessarily reflect the views, policies, or endorsements, either expressed or implied, of NIH, ONR, or the U.S. Government. 

\bibliographystyle{abbrvnat}
\bibliography{_bibliography/main}

\begin{thebibliography}{59}
\providecommand{\natexlab}[1]{#1}
\providecommand{\url}[1]{\texttt{#1}}
\expandafter\ifx\csname urlstyle\endcsname\relax
  \providecommand{\doi}[1]{doi: #1}\else
  \providecommand{\doi}{doi: \begingroup \urlstyle{rm}\Url}\fi

\bibitem[Arora et~al.(2022)Arora, Narayan, Chen, Orr, Guha, Bhatia, Chami,
  Sala, and R{\'e}]{arora2022ask}
S.~Arora, A.~Narayan, M.~F. Chen, L.~J. Orr, N.~Guha, K.~Bhatia, I.~Chami,
  F.~Sala, and C.~R{\'e}.
\newblock Ask me anything: A simple strategy for prompting language models.
\newblock \emph{arXiv preprint arXiv:2210.02441}, 2022.

\bibitem[Basri et~al.(2020)Basri, Galun, Geifman, Jacobs, Kasten, and
  Kritchman]{basri2020frequency}
R.~Basri, M.~Galun, A.~Geifman, D.~Jacobs, Y.~Kasten, and S.~Kritchman.
\newblock Frequency bias in neural networks for input of non-uniform density.
\newblock In \emph{International Conference on Machine Learning}, pages
  685--694. PMLR, 2020.

\bibitem[Black et~al.(2021)Black, Gao, Wang, Leahy, and Biderman]{gpt-neo}
S.~Black, L.~Gao, P.~Wang, C.~Leahy, and S.~Biderman.
\newblock {GPT-Neo: Large Scale Autoregressive Language Modeling with
  Mesh-Tensorflow}, Mar. 2021.
\newblock URL \url{https://doi.org/10.5281/zenodo.5297715}.
\newblock {If you use this software, please cite it using these metadata.}

\bibitem[Brown et~al.(2020)Brown, Mann, Ryder, Subbiah, Kaplan, Dhariwal,
  Neelakantan, Shyam, Sastry, Askell, et~al.]{brown2020language}
T.~Brown, B.~Mann, N.~Ryder, M.~Subbiah, J.~D. Kaplan, P.~Dhariwal,
  A.~Neelakantan, P.~Shyam, G.~Sastry, A.~Askell, et~al.
\newblock Language models are few-shot learners.
\newblock \emph{Advances in neural information processing systems},
  33:\penalty0 1877--1901, 2020.

\bibitem[Chen(1984)]{chen1984linear}
C.-T. Chen.
\newblock \emph{Linear system theory and design}.
\newblock Saunders college publishing, 1984.

\bibitem[Child et~al.(2019)Child, Gray, Radford, and
  Sutskever]{child2019generating}
R.~Child, S.~Gray, A.~Radford, and I.~Sutskever.
\newblock Generating long sequences with sparse transformers.
\newblock \emph{arXiv preprint arXiv:1904.10509}, 2019.

\bibitem[Cramer(2021)]{cramer2021alphafold2}
P.~Cramer.
\newblock Alphafold2 and the future of structural biology.
\newblock \emph{Nature structural \& molecular biology}, 28\penalty0
  (9):\penalty0 704--705, 2021.

\bibitem[Cubuk et~al.(2020)Cubuk, Zoph, Shlens, and Le]{cubuk2020randaugment}
E.~D. Cubuk, B.~Zoph, J.~Shlens, and Q.~V. Le.
\newblock Randaugment: Practical automated data augmentation with a reduced
  search space.
\newblock In \emph{Proceedings of the IEEE/CVF conference on computer vision
  and pattern recognition workshops}, pages 702--703, 2020.

\bibitem[Dao et~al.(2019)Dao, Gu, Eichhorn, Rudra, and R{\'e}]{dao2019learning}
T.~Dao, A.~Gu, M.~Eichhorn, A.~Rudra, and C.~R{\'e}.
\newblock Learning fast algorithms for linear transforms using butterfly
  factorizations.
\newblock In \emph{International conference on machine learning}, pages
  1517--1527. PMLR, 2019.

\bibitem[Dao et~al.(2022{\natexlab{a}})Dao, Chen, Sohoni, Desai, Poli, Grogan,
  Liu, Rao, Rudra, and R{\'e}]{dao2022monarch}
T.~Dao, B.~Chen, N.~S. Sohoni, A.~Desai, M.~Poli, J.~Grogan, A.~Liu, A.~Rao,
  A.~Rudra, and C.~R{\'e}.
\newblock Monarch: Expressive structured matrices for efficient and accurate
  training.
\newblock In \emph{International Conference on Machine Learning}, pages
  4690--4721. PMLR, 2022{\natexlab{a}}.

\bibitem[Dao et~al.(2022{\natexlab{b}})Dao, Fu, Ermon, Rudra, and
  R{\'e}]{dao2022flashattention}
T.~Dao, D.~Y. Fu, S.~Ermon, A.~Rudra, and C.~R{\'e}.
\newblock Flashattention: Fast and memory-efficient exact attention with
  io-awareness.
\newblock \emph{arXiv preprint arXiv:2205.14135}, 2022{\natexlab{b}}.

\bibitem[Dao et~al.(2022{\natexlab{c}})Dao, Fu, Saab, Thomas, Rudra, and
  R{\'e}]{dao2022hungry}
T.~Dao, D.~Y. Fu, K.~K. Saab, A.~W. Thomas, A.~Rudra, and C.~R{\'e}.
\newblock Hungry hungry hippos: Towards language modeling with state space
  models.
\newblock \emph{arXiv preprint arXiv:2212.14052}, 2022{\natexlab{c}}.

\bibitem[Dosovitskiy et~al.(2020)Dosovitskiy, Beyer, Kolesnikov, Weissenborn,
  Zhai, Unterthiner, Dehghani, Minderer, Heigold, Gelly,
  et~al.]{dosovitskiy2020image}
A.~Dosovitskiy, L.~Beyer, A.~Kolesnikov, D.~Weissenborn, X.~Zhai,
  T.~Unterthiner, M.~Dehghani, M.~Minderer, G.~Heigold, S.~Gelly, et~al.
\newblock An image is worth 16x16 words: Transformers for image recognition at
  scale.
\newblock \emph{arXiv preprint arXiv:2010.11929}, 2020.

\bibitem[Elhage et~al.(2021)Elhage, Nanda, Olsson, Henighan, Joseph, Mann,
  Askell, Bai, Chen, Conerly, et~al.]{elhage2021mathematical}
N.~Elhage, N.~Nanda, C.~Olsson, T.~Henighan, N.~Joseph, B.~Mann, A.~Askell,
  Y.~Bai, A.~Chen, T.~Conerly, et~al.
\newblock A mathematical framework for transformer circuits.
\newblock \emph{Transformer Circuits Thread}, 2021.

\bibitem[Fukushima and Miyake(1982)]{fukushima1982neocognitron}
K.~Fukushima and S.~Miyake.
\newblock Neocognitron: A self-organizing neural network model for a mechanism
  of visual pattern recognition.
\newblock In \emph{Competition and cooperation in neural nets}, pages 267--285.
  Springer, 1982.

\bibitem[Gao et~al.(2020)Gao, Biderman, Black, Golding, Hoppe, Foster, Phang,
  He, Thite, Nabeshima, et~al.]{gao2020pile}
L.~Gao, S.~Biderman, S.~Black, L.~Golding, T.~Hoppe, C.~Foster, J.~Phang,
  H.~He, A.~Thite, N.~Nabeshima, et~al.
\newblock The pile: An 800gb dataset of diverse text for language modeling.
\newblock \emph{arXiv preprint arXiv:2101.00027}, 2020.

\bibitem[Garg et~al.(2022)Garg, Tsipras, Liang, and Valiant]{garg2022can}
S.~Garg, D.~Tsipras, P.~Liang, and G.~Valiant.
\newblock What can transformers learn in-context? a case study of simple
  function classes.
\newblock \emph{arXiv preprint arXiv:2208.01066}, 2022.

\bibitem[Gu et~al.(2020)Gu, Dao, Ermon, Rudra, and R{\'e}]{gu2020hippo}
A.~Gu, T.~Dao, S.~Ermon, A.~Rudra, and C.~R{\'e}.
\newblock Hippo: Recurrent memory with optimal polynomial projections.
\newblock \emph{Advances in Neural Information Processing Systems},
  33:\penalty0 1474--1487, 2020.

\bibitem[Gu et~al.(2021)Gu, Goel, and R{\'e}]{gu2021efficiently}
A.~Gu, K.~Goel, and C.~R{\'e}.
\newblock Efficiently modeling long sequences with structured state spaces.
\newblock \emph{arXiv preprint arXiv:2111.00396}, 2021.

\bibitem[He et~al.(2016)He, Zhang, Ren, and Sun]{he2016deep}
K.~He, X.~Zhang, S.~Ren, and J.~Sun.
\newblock Deep residual learning for image recognition.
\newblock In \emph{Proceedings of the IEEE conference on computer vision and
  pattern recognition}, pages 770--778, 2016.

\bibitem[Hendrycks et~al.(2019)Hendrycks, Mu, Cubuk, Zoph, Gilmer, and
  Lakshminarayanan]{hendrycks2019augmix}
D.~Hendrycks, N.~Mu, E.~D. Cubuk, B.~Zoph, J.~Gilmer, and B.~Lakshminarayanan.
\newblock Augmix: A simple data processing method to improve robustness and
  uncertainty.
\newblock \emph{arXiv preprint arXiv:1912.02781}, 2019.

\bibitem[Hoffmann et~al.(2022)Hoffmann, Borgeaud, Mensch, Buchatskaya, Cai,
  Rutherford, Casas, Hendricks, Welbl, Clark, et~al.]{hoffmann2022training}
J.~Hoffmann, S.~Borgeaud, A.~Mensch, E.~Buchatskaya, T.~Cai, E.~Rutherford,
  D.~d.~L. Casas, L.~A. Hendricks, J.~Welbl, A.~Clark, et~al.
\newblock Training compute-optimal large language models.
\newblock \emph{arXiv preprint arXiv:2203.15556}, 2022.

\bibitem[Huang et~al.(2016)Huang, Sun, Liu, Sedra, and
  Weinberger]{huang2016deep}
G.~Huang, Y.~Sun, Z.~Liu, D.~Sedra, and K.~Q. Weinberger.
\newblock Deep networks with stochastic depth.
\newblock In \emph{European conference on computer vision}, pages 646--661.
  Springer, 2016.

\bibitem[Kitaev et~al.(2020)Kitaev, Kaiser, and Levskaya]{kitaev2020reformer}
N.~Kitaev, {\L}.~Kaiser, and A.~Levskaya.
\newblock Reformer: The efficient transformer.
\newblock \emph{arXiv preprint arXiv:2001.04451}, 2020.

\bibitem[LeCun et~al.(1998)LeCun, Bottou, Bengio, and
  Haffner]{lecun1998gradient}
Y.~LeCun, L.~Bottou, Y.~Bengio, and P.~Haffner.
\newblock Gradient-based learning applied to document recognition.
\newblock \emph{Proceedings of the IEEE}, 86\penalty0 (11):\penalty0
  2278--2324, 1998.

\bibitem[Li et~al.(2015)Li, Yang, Martin, Ho, and Ying]{li2015butterfly}
Y.~Li, H.~Yang, E.~R. Martin, K.~L. Ho, and L.~Ying.
\newblock Butterfly factorization.
\newblock \emph{Multiscale Modeling \& Simulation}, 13\penalty0 (2):\penalty0
  714--732, 2015.

\bibitem[Li et~al.(2022)Li, Cai, Zhang, Chen, and Dey]{li2022makes}
Y.~Li, T.~Cai, Y.~Zhang, D.~Chen, and D.~Dey.
\newblock What makes convolutional models great on long sequence modeling?
\newblock \emph{arXiv preprint arXiv:2210.09298}, 2022.

\bibitem[Li et~al.(2020)Li, Kovachki, Azizzadenesheli, Liu, Bhattacharya,
  Stuart, and Anandkumar]{li2020fourier}
Z.~Li, N.~Kovachki, K.~Azizzadenesheli, B.~Liu, K.~Bhattacharya, A.~Stuart, and
  A.~Anandkumar.
\newblock Fourier neural operator for parametric partial differential
  equations.
\newblock \emph{arXiv preprint arXiv:2010.08895}, 2020.

\bibitem[Liang et~al.(2022)Liang, Bommasani, Lee, Tsipras, Soylu, Yasunaga,
  Zhang, Narayanan, Wu, Kumar, et~al.]{liang2022holistic}
P.~Liang, R.~Bommasani, T.~Lee, D.~Tsipras, D.~Soylu, M.~Yasunaga, Y.~Zhang,
  D.~Narayanan, Y.~Wu, A.~Kumar, et~al.
\newblock Holistic evaluation of language models.
\newblock \emph{arXiv preprint arXiv:2211.09110}, 2022.

\bibitem[Massaroli et~al.(2020)Massaroli, Poli, Park, Yamashita, and
  Asama]{massaroli2020dissecting}
S.~Massaroli, M.~Poli, J.~Park, A.~Yamashita, and H.~Asama.
\newblock Dissecting neural odes.
\newblock \emph{Advances in Neural Information Processing Systems},
  33:\penalty0 3952--3963, 2020.

\bibitem[Mehta et~al.(2022)Mehta, Gupta, Cutkosky, and
  Neyshabur]{mehta2022long}
H.~Mehta, A.~Gupta, A.~Cutkosky, and B.~Neyshabur.
\newblock Long range language modeling via gated state spaces.
\newblock \emph{arXiv preprint arXiv:2206.13947}, 2022.

\bibitem[Mildenhall et~al.(2021)Mildenhall, Srinivasan, Tancik, Barron,
  Ramamoorthi, and Ng]{mildenhall2021nerf}
B.~Mildenhall, P.~P. Srinivasan, M.~Tancik, J.~T. Barron, R.~Ramamoorthi, and
  R.~Ng.
\newblock Nerf: Representing scenes as neural radiance fields for view
  synthesis.
\newblock \emph{Communications of the ACM}, 65\penalty0 (1):\penalty0 99--106,
  2021.

\bibitem[Nguyen et~al.(2022)Nguyen, Goel, Gu, Downs, Shah, Dao, Baccus, and
  R{\'e}]{nguyen2022s4nd}
E.~Nguyen, K.~Goel, A.~Gu, G.~W. Downs, P.~Shah, T.~Dao, S.~A. Baccus, and
  C.~R{\'e}.
\newblock S4nd: Modeling images and videos as multidimensional signals using
  state spaces.
\newblock \emph{arXiv preprint arXiv:2210.06583}, 2022.

\bibitem[Olsson et~al.(2022)Olsson, Elhage, Nanda, Joseph, DasSarma, Henighan,
  Mann, Askell, Bai, Chen, et~al.]{olsson2022context}
C.~Olsson, N.~Elhage, N.~Nanda, N.~Joseph, N.~DasSarma, T.~Henighan, B.~Mann,
  A.~Askell, Y.~Bai, A.~Chen, et~al.
\newblock In-context learning and induction heads.
\newblock \emph{arXiv preprint arXiv:2209.11895}, 2022.

\bibitem[Oppenheim et~al.(1997)Oppenheim, Willsky, Nawab, and
  Ding]{oppenheim1997signals}
A.~V. Oppenheim, A.~S. Willsky, S.~H. Nawab, and J.-J. Ding.
\newblock \emph{Signals and systems}, volume~2.
\newblock Prentice hall Upper Saddle River, NJ, 1997.

\bibitem[Paperno et~al.(2016)Paperno, Kruszewski, Lazaridou, Pham, Bernardi,
  Pezzelle, Baroni, Boleda, and Fern{\'a}ndez]{paperno2016lambada}
D.~Paperno, G.~Kruszewski, A.~Lazaridou, Q.~N. Pham, R.~Bernardi, S.~Pezzelle,
  M.~Baroni, G.~Boleda, and R.~Fern{\'a}ndez.
\newblock The lambada dataset: Word prediction requiring a broad discourse
  context.
\newblock \emph{arXiv preprint arXiv:1606.06031}, 2016.

\bibitem[Peng(2021)]{PENG_RWKV-LM_2021}
B.~Peng.
\newblock {RWKV-LM}, 8 2021.
\newblock URL \url{https://github.com/BlinkDL/RWKV-LM}.

\bibitem[Polyak and Juditsky(1992)]{polyak1992ema}
B.~T. Polyak and A.~B. Juditsky.
\newblock Acceleration of stochastic approximation by averaging.
\newblock \emph{SIAM journal on control and optimization}, 30\penalty0
  (4):\penalty0 838--855, 1992.

\bibitem[Power et~al.(2022)Power, Burda, Edwards, Babuschkin, and
  Misra]{power2022grokking}
A.~Power, Y.~Burda, H.~Edwards, I.~Babuschkin, and V.~Misra.
\newblock Grokking: Generalization beyond overfitting on small algorithmic
  datasets.
\newblock \emph{arXiv preprint arXiv:2201.02177}, 2022.

\bibitem[Radford et~al.(2022)Radford, Kim, Xu, Brockman, McLeavey, and
  Sutskever]{radford2022robust}
A.~Radford, J.~W. Kim, T.~Xu, G.~Brockman, C.~McLeavey, and I.~Sutskever.
\newblock Robust speech recognition via large-scale weak supervision.
\newblock \emph{arXiv preprint arXiv:2212.04356}, 2022.

\bibitem[Rae et~al.(2019)Rae, Potapenko, Jayakumar, Hillier, and
  Lillicrap]{raecompressive2019}
J.~W. Rae, A.~Potapenko, S.~M. Jayakumar, C.~Hillier, and T.~P. Lillicrap.
\newblock Compressive transformers for long-range sequence modelling.
\newblock \emph{arXiv preprint}, 2019.
\newblock URL \url{https://arxiv.org/abs/1911.05507}.

\bibitem[Romero et~al.(2021{\natexlab{a}})Romero, Bruintjes, Tomczak, Bekkers,
  Hoogendoorn, and van Gemert]{romero2021flexconv}
D.~W. Romero, R.-J. Bruintjes, J.~M. Tomczak, E.~J. Bekkers, M.~Hoogendoorn,
  and J.~C. van Gemert.
\newblock Flexconv: Continuous kernel convolutions with differentiable kernel
  sizes.
\newblock \emph{arXiv preprint arXiv:2110.08059}, 2021{\natexlab{a}}.

\bibitem[Romero et~al.(2021{\natexlab{b}})Romero, Kuzina, Bekkers, Tomczak, and
  Hoogendoorn]{romero2021ckconv}
D.~W. Romero, A.~Kuzina, E.~J. Bekkers, J.~M. Tomczak, and M.~Hoogendoorn.
\newblock Ckconv: Continuous kernel convolution for sequential data.
\newblock \emph{arXiv preprint arXiv:2102.02611}, 2021{\natexlab{b}}.

\bibitem[Ronneberger et~al.(2015)Ronneberger, Fischer, and
  Brox]{ronneberger2015u}
O.~Ronneberger, P.~Fischer, and T.~Brox.
\newblock U-net: Convolutional networks for biomedical image segmentation.
\newblock In \emph{International Conference on Medical image computing and
  computer-assisted intervention}, pages 234--241. Springer, 2015.

\bibitem[Roy et~al.(2021)Roy, Saffar, Vaswani, and Grangier]{roy2021efficient}
A.~Roy, M.~Saffar, A.~Vaswani, and D.~Grangier.
\newblock Efficient content-based sparse attention with routing transformers.
\newblock \emph{Transactions of the Association for Computational Linguistics},
  9:\penalty0 53--68, 2021.

\bibitem[Schlag et~al.(2021)Schlag, Irie, and Schmidhuber]{schlag2021linear}
I.~Schlag, K.~Irie, and J.~Schmidhuber.
\newblock Linear transformers are secretly fast weight programmers.
\newblock In \emph{International Conference on Machine Learning}, pages
  9355--9366. PMLR, 2021.

\bibitem[Selesnick and Burrus(2017)]{selesnick2017fast}
I.~W. Selesnick and C.~S. Burrus.
\newblock Fast convolution and filtering.
\newblock In \emph{The Digital Signal Processing Handbook}, pages 8--1. CRC
  Press, 2017.

\bibitem[Sitzmann et~al.(2020)Sitzmann, Martel, Bergman, Lindell, and
  Wetzstein]{sitzmann2020implicit}
V.~Sitzmann, J.~N. Martel, A.~W. Bergman, D.~B. Lindell, and G.~Wetzstein.
\newblock Implicit neural representations with periodic activation functions.
\newblock \emph{arXiv preprint arXiv:2006.09661}, 2020.

\bibitem[Szegedy et~al.(2016)Szegedy, Vanhoucke, Ioffe, Shlens, and
  Wojna]{szegedy2016rethinking}
C.~Szegedy, V.~Vanhoucke, S.~Ioffe, J.~Shlens, and Z.~Wojna.
\newblock Rethinking the inception architecture for computer vision.
\newblock In \emph{Proceedings of the IEEE conference on computer vision and
  pattern recognition}, pages 2818--2826, 2016.

\bibitem[Tu et~al.(2022)Tu, Talebi, Zhang, Yang, Milanfar, Bovik, and
  Li]{tu2022maxvit}
Z.~Tu, H.~Talebi, H.~Zhang, F.~Yang, P.~Milanfar, A.~Bovik, and Y.~Li.
\newblock Maxvit: Multi-axis vision transformer.
\newblock \emph{arXiv preprint arXiv:2204.01697}, 2022.

\bibitem[Vaswani et~al.(2017)Vaswani, Shazeer, Parmar, Uszkoreit, Jones, Gomez,
  Kaiser, and Polosukhin]{vaswani2017attention}
A.~Vaswani, N.~Shazeer, N.~Parmar, J.~Uszkoreit, L.~Jones, A.~N. Gomez,
  {\L}.~Kaiser, and I.~Polosukhin.
\newblock Attention is all you need.
\newblock In \emph{Advances in neural information processing systems}, pages
  5998--6008, 2017.

\bibitem[Wang et~al.(2019)Wang, Pruksachatkun, Nangia, Singh, Michael, Hill,
  Levy, and Bowman]{wang2019superglue}
A.~Wang, Y.~Pruksachatkun, N.~Nangia, A.~Singh, J.~Michael, F.~Hill, O.~Levy,
  and S.~Bowman.
\newblock Superglue: A stickier benchmark for general-purpose language
  understanding systems.
\newblock \emph{Advances in neural information processing systems}, 32, 2019.

\bibitem[Wang et~al.(2020)Wang, Li, Khabsa, Fang, and Ma]{wang2020linformer}
S.~Wang, B.~Z. Li, M.~Khabsa, H.~Fang, and H.~Ma.
\newblock Linformer: Self-attention with linear complexity.
\newblock \emph{arXiv preprint arXiv:2006.04768}, 2020.

\bibitem[Yuan et~al.(2021)Yuan, Chen, Wang, Yu, Shi, Jiang, Tay, Feng, and
  Yan]{yuan2021tokens}
L.~Yuan, Y.~Chen, T.~Wang, W.~Yu, Y.~Shi, Z.-H. Jiang, F.~E. Tay, J.~Feng, and
  S.~Yan.
\newblock Tokens-to-token vit: Training vision transformers from scratch on
  imagenet.
\newblock In \emph{Proceedings of the IEEE/CVF international conference on
  computer vision}, pages 558--567, 2021.

\bibitem[Yun et~al.(2019)Yun, Han, Oh, Chun, Choe, and Yoo]{yun2019cutmix}
S.~Yun, D.~Han, S.~J. Oh, S.~Chun, J.~Choe, and Y.~Yoo.
\newblock Cutmix: Regularization strategy to train strong classifiers with
  localizable features.
\newblock In \emph{Proceedings of the IEEE/CVF international conference on
  computer vision}, pages 6023--6032, 2019.

\bibitem[Zhai et~al.(2021)Zhai, Talbott, Srivastava, Huang, Goh, Zhang, and
  Susskind]{zhai2021attention}
S.~Zhai, W.~Talbott, N.~Srivastava, C.~Huang, H.~Goh, R.~Zhang, and
  J.~Susskind.
\newblock An attention free transformer.
\newblock \emph{arXiv preprint arXiv:2105.14103}, 2021.

\bibitem[Zhang et~al.(2017)Zhang, Cisse, Dauphin, and
  Lopez-Paz]{zhang2017mixup}
H.~Zhang, M.~Cisse, Y.~N. Dauphin, and D.~Lopez-Paz.
\newblock mixup: Beyond empirical risk minimization.
\newblock \emph{arXiv preprint arXiv:1710.09412}, 2017.

\bibitem[Zhang et~al.(2022)Zhang, Backurs, Bubeck, Eldan, Gunasekar, and
  Wagner]{zhang2022unveiling}
Y.~Zhang, A.~Backurs, S.~Bubeck, R.~Eldan, S.~Gunasekar, and T.~Wagner.
\newblock Unveiling transformers with lego: a synthetic reasoning task.
\newblock \emph{arXiv preprint arXiv:2206.04301}, 2022.

\bibitem[Zhong et~al.(2020)Zhong, Zheng, Kang, Li, and Yang]{zhong2020random}
Z.~Zhong, L.~Zheng, G.~Kang, S.~Li, and Y.~Yang.
\newblock Random erasing data augmentation.
\newblock In \emph{Proceedings of the AAAI conference on artificial
  intelligence}, volume~34, pages 13001--13008, 2020.

\end{thebibliography}
\clearpage
\appendix
\rule[0pt]{\columnwidth}{1pt}
\begin{center}
    \huge{Hyena Hierarchy} \\
    \vspace{0.3cm}
    \emph{Supplementary Material}
\end{center}
\rule[0pt]{\columnwidth}{1.5pt}
\doparttoc
\tableofcontents
\clearpage
\section{Experimental Details}
\label{appendix:experiment-details}
An implementation of ${\sf Hyena}$ can be found at \href{https://github.com/HazyResearch/safari}{this link}.

\subsection{Mechanistic Design Synthetic Benchmarks}\label{app:icl}
Our synthetic reasoning are inspired by mechanistic interpretability \citep{elhage2021mathematical}, \textit{in-context learning} (ICL) \citep{garg2022can} and language model benchmarking \citep{liang2022holistic} research. The evaluation revolves around $4$ main tasks:
\begin{itemize}[leftmargin=0.2in]
    \item \textbf{Associative recall:} Each string is produced by concatenating key-value tuples from a different random dictionary. This test verifies whether a model is able to extract right value given a key as prompt, effectively applying a data-controlled shift (delay).
    \item \textbf{Majority voting and counting:} Testing if a model can \textit{densely} activate its data-controlled matrix i.e., through many non-zero entries (consider the string '\textit{a a a a a a a a a a b} $\rightarrow$ \textit{a}').
    \item \textbf{ICL of linear functions:} Verifying whether a model can perform ICL on real-valued inputs. Prompts are generated as $x_1, w^k x_1, \dots, x_n \rightarrow w^k x_n$, where both $x_k$ and $w^k \in R^{n_o}$ are sampled from a normal distribution.  
    \item \textbf{Arithmetic:} Basic capability check. 
\end{itemize}

For each task, we train models using the hyperparameters shown in Table \ref{tab:synthetics}. We consider increasing settings of difficulty controlled by sequence length, spanning values $1024, 2048, 4098, 8196, 16392, 32784, 65568, 131136$ and vocabulary sizes $10, 20, 30, 40$. For ICL of functions, we vary instead the dimension $n_o$.

Note that for associative recall on longer sequences, multiple copies of key-value tuples appear in the prompt. To see this, consider how likely it is to sample multiple copies of a particular key-value pair with a vocabulary size of $40$, in order to form a sequence of $100$k characters. Models capable of looking further back in the sequence effectively see more data, and can solve challenging versions of the in-context learning task. Increasing the vocabulary size has the increasing the average distance between instances of the same key-value pair in each prompt, highlighting performance gaps between different approaches.

\begin{table}[ht]
      \small
      \caption{{(\bf Hyperparameter settings for reasoning and in-context learning tasks.)}.
      }
        \centering
        \begin{tabular}{lcc}
            \toprule
            Optimizer & AdamW \\
            Optimizer momentum & $\beta_1,\beta_2=0.9,0.98$ \\
            Base learning rate & 0.0005 \\
            Weight decay & 0.1 \\
            Dropout & None \\
            Batch size & 32 \\\
            Training epochs & 200 \\
            Num samples & 2000 \\ 
            Learning rate schedule & cosine decay \\
            Warmup epochs & 10 \\
            Warmup schedule & linear \\
            Number of layers & 2 \\ 
            Width & 64 \\
            \bottomrule
        \end{tabular}
        \label{tab:synthetics}
\end{table}

\paragraph{Long convolution comparisons:}
We compare different convolution parametrizations, embedding them in an order $2$ ${\sf Hyena}$ operator. All convolutions are applied separately to input channels (referred to as single-input single-output (SISO) in signal processing, or \textit{depthwise} in other machine learning contexts).

\begin{itemize}[leftmargin=0.1in]
    \item Conv1d: Explicit convolutions (regular convolution layers with fixed filter size). We use a fixed filter size of $64$, to match parameters of the other approaches.
    \item FNO: Filters parametrized explicitly in the frequency-domain \citep{li2020fourier}. We set the number of modes to $64$.
    \item H3: Implicit parametrization using state-space models (SSMs), and in particular the standard S4 \citep{gu2021efficiently}. We set the state dimension to $64$.
    \item TransferFunc: Implicit parametrization via transfer functions, a classical system-theoretic generalization of SSMs. Transfer functions are defined by a ratio of polynomials (we parametrize the coefficients, and evaluate the polynomials efficiently via FFTs). We set the order to $64$.
    \item CKConv: Implicit parametrization using {$\sf FFN$s} \citep{romero2021ckconv}. 
    \item item ${\sf Hyena}$: Combination of implicit parametrizations via {$\sf FFN$s} (with exponential decay modulation as shown in Figure \ref{fig:modul}), and short explicit filters.
\end{itemize}

CKConv and ${\sf Hyena}$ use the same size of ${\sf FFNs}$ (width $32$ to match in parameters).  
    
In Table \ref{synthetic2}, we report additional results on the challenging setting of sequence length $131072$ and vocabulary size $30$. Implicit parametrizations of convolutions outperform explicit parametrizations on associative recall, with CKConv and ${\sf Hyena}$ greatly improving on the ability to extract the right key, value relations from different inputs. In Appendix \ref{app:add_results}, we discuss how results on our synthetic tasks can be indicative of performance at a larger scale. 

\begin{table}[h]
\label{synthetic2}
\small
\centering
\vspace{-2mm}
\caption{Test accuracy (\%) in associative recall on sequences of length $131072$, vocabulary size $30$.}
\vspace{2mm}
\resizebox{0.45\linewidth}{!}
{
\setlength{\tabcolsep}{4pt}
\vspace{3em}
\begin{tabular}{@{}c|ccccccc@{}}
\toprule
${\sf Hyena}$ &\multicolumn{1}{c}{CKConv}&\multicolumn{1}{c}{TransferFunc} & \multicolumn{1}{c}{H3} & \multicolumn{1}{c} {FNO} & \multicolumn{1}{c}{Conv1d} \\
\midrule 
97.2 & 14.3 & 0.5 & 0.6 & 0.3 & 0.5 \\
\bottomrule
\end{tabular}
}
\end{table}

\paragraph{Operator comparisons:}
We compare different models on the same associative recall task, using hyperparameters in Table \ref{tab:synthetics}. ${\sf Hyena}$ uses our filter parametrization with decay windowing for long convolutions, and short explicit convolutions of size $3$ after the dense input projections. All other models use defaults from their largest scale experiment, while keeping the size to $2$ layers and width $64$.

\paragraph{A note on Transformer performance}
Transformers can solve associative recall tasks with longer sequences, provided the length does not prevent them from fitting in memory, and enough examples are present in the training data. In all our experiments, we keep the number of samples fixed ($2000$), a regime where Transformers struggle to find the generalizing solution (see Table \ref{synthetic2}).

For shorter sequences (see Appendix \ref{app:add_results}), Transformers solve the task easily even with limited data, comparably to {\sf Hyena}. 

More broadly, these different properties of attention and attention-free token-mixing layers may explain improved performance when they are combined in hybrid architectures \citep{dao2022hungry}. The focus on this work has been identifying an architecture capable of performing without attention, which is necessary to tackle domains where long sequences are common. However, when training with shorter sequences (up to $8$k), if final downstream performance is the only metric of interest, improved results can be obtained by hybridizing our models similarly to H3 \citep{dao2022hungry}.

\subsection{Language Modeling}
\paragraph{WikiText103:}
We train $125$M parameter models on {\sc WikiText103} and compare perplexity to Transformers, hybrid models such as H3 \citep{dao2022hungry}, and other variants of subquadratic attention. All models use the same GPT2 tokenizer with vocabulary size $50257$. We test order $3$ ${\sf Hyena}$ with our proposed filter parametrization for two long convolutions, and a shorter explicit convolution on the third. We also consider ${\sf Hyena}$ (slim) that are $1.5$x deeper than Transformers ($12$ versus $18$ layers), with width multiplier of the FFNs set to $2$. We find trading-off width for depth to be generally favourable. These modifications are made possible by the reduction in overall FLOPs of {\sf Hyena} operators compared to self-attention, in particular non-parametric FLOPs which include materialization of the attention matrix, application of softmax, and matrix-value reduction. 

\begin{table}[ht]
      \small
      \caption{Hyperparameter settings for {\sc The Pile}, $125$M).
      }
        \centering
        \begin{tabular}{lcc}
            \toprule
            Optimizer & AdamW \\
            Optimizer momentum & $\beta_1,\beta_2=0.9,0.98$ \\
            Peak learning rate & 0.0006 \\
            Warmup learning rate init & 0.000001 \\
            Learning rate min &  0.00006 \\ 
            Weight decay & 0.1 \\
            Dropout & None \\
            Batch size & 256 \\
            Learning rate schedule & cosine decay \\
            Warmup schedule & linear \\
            \bottomrule
        \end{tabular}
        \label{tab:pilehyper}
\end{table}

\paragraph{The Pile:}
We follow a same procedure and train $125$M and $355$M-sized models on {\sc The Pile} \citep{gao2020pile}. Hyperparameters are reported in Table \ref{tab:pilehyper}. Hyperparameters for $355$M are the same beyond a reduction in peak learning rate to $4\cdot 10^{-4}$. For larger models ($1.3$B), we set a learning rate of $2.2 \cdot 10^{-4}$.

We perform three experiments for each model type and size, and train for $5, 10, 15$ billion tokens at a sequence length $2024$ and global batch size $256$. All models are trained on a single node of $8$ $A100$ $80$GB GPUs. We use order $2$ ${\sf Hyena}$s, with the same architectural considerations described above for {\sc WikiText103}.
In addition to our data scaling experiments at $5$, $10$ and $15$ billion tokens, we provide preliminary results for models at the $1.3$B parameter scale ($10.8$ perplexity after $5$ billion tokens), and train a $153$M model ($130$ billion tokens), reaching a perplexity of $9.8$. The $153$M is the same used in our downstream evaluation on SuperGLUE. 

Training hyperparameters match those of standard GPT training pipelines, and are thus likely suboptimal for new attention-free architectures such as {\sf Hyena}. We run some preliminary experiments and find that e.g., some modifications to the learning rate schedule, currently involving linear warmup and cosine decay, to improve perplexity at convergence of {\sf Hyena} models (we recommend slightly lower learning rates for Hyena models compared to GPT of a similar size). Despite these findings, we use standard GPT hyperparameters for both GPT and {\sf Hyena}.
\paragraph{PG-19}

We also report results of additional training runs on other datasets. We train a ${\sf Hyena}$ $153$M model on the standard PG-19 long-range corpus \citep{raecompressive2019}, with a context length of $16$k tokens, reaching a test perplexity of $14.6$ (using the standard GPT2 tokenizer) in $8$ epochs.
\paragraph{Architectures}
Architectural hyperparameters for {\sf Hyena} are shown in Table \ref{hyena_arch}. We use sine as an activation function for the {\sf FFN} of {\sf Hyena} filters. 

\begin{table}[!bh]
\small
\centering
\caption{{\sf Hyena} architecture hyperparameters.}
\label{hyena_arch}
\setlength{\tabcolsep}{4pt}
\begin{tabular}{@{}c|cccccc@{}}
\toprule
Size & depth & width & {\sf FFN} width & filter {\sf FFN} width & filter {\sf FFN} depth & sine freq. \\
\midrule 
$125$M & $12$ & $768$ & $3072$ & $64$ & $4$ & $14$ \\ 
$125$M-slim &$18$ & $768$ & $1536$ & $64$ & $4$ & $14$\\ 
$153$M & $18$ & $864$ & $1728$ & $64$  & $4$ & $14$\\  
$355$M & $36$ & $1024$ & $2048$ & $64$ & $4$ & $14$ \\ 
$1.3$B & $36$ & $2048$ & $4096$ & $64$ & $4$ & $14$ \\ 
\bottomrule
\end{tabular}
\end{table}

\paragraph{FLOP computation}
The number of \textit{floating point operations} (FLOPs) reported in the main text are computed using the same strategy as in \citep{hoffmann2022training}. For GPT, we do not use the approximation, opting instead for the more accurate formula based on FLOP counts of individual layers. In the case of {\sf Hyena}, FLOPs are computed using the same method, except attention layers are replaced by:
\begin{itemize}
    \item[i.] Projections: order $\times$ d\_model $\times$ d\_model $\times$ seq\_len.
    \item[ii.] Short conv on projections: order $\times$ d\_model $\times$ seq\_len $\times$ filter\_len (usually $3$).
    \item[iii.] FFTConv: $5$ $\times$ (order - 1) $\times$ d\_model $\times$ $\log(\text{seq\_len})$
    $\times$ seq\_len.
    \item[iv.] Output: d\_model $\times$ d\_model $\times$ seq\_len.
\end{itemize}
with a leading factor $2$ to account for both additions and multiplications. 

\subsection{Downstream Evaluation}
\paragraph{SuperGLUE:} We evaluate models on the SuperGLUE \citep{wang2019superglue} with the parsing pipeline of \citep{arora2022ask}. For all tasks except WIC, CB and BoolQ, we generate a response using greedy decoding, then check for the gold label. WIC, CB and BoolQ use logit scoring instead of generation. 

\paragraph{Models}
The models considered are the open-source checkpoint of GPTNeo $125$M trained for $300$B tokens {\sc The Pile}, and the RWKV-v4 $169$M checkpoint trained for $332$B tokens on {\sc The Pile}. {\sf Hyena} is a $153$M model trained for $137$B tokens on {\sc The Pile}.
\paragraph{LAMBADA:} We evaluate ${\sf Hyena}$ on the LAMBADA \citep{paperno2016lambada} task. We apply a stop word filter and check whether predictions for all tokens corresponding to the last word agree with the ground truth. The small ${\sf Hyena}$ model trained on $137$B tokens reaches $44.64\%$ accuracy. 

\subsection{Image Classification}
\label{appendix:image-classification}

\paragraph
a{ImageNet:} We use ImageNet-1k which consists of 1000 classes and 1.3M images and train from scratch with no outside data on 8 Nvidia A100 GPUs. In our ViT benchmark, we swap the attention layers with the Hyena operator defined in our language experiments, and remove the class token and positional embeddings, similar to S4ND \citep{nguyen2022s4nd}. The parameter count is kept similar at 87M ViT-B (base) vs 88M Hyena-ViT. The training procedure from T2T-ViT \citep{yuan2021tokens} is used, including augmentations such as RandAugment \citep{cubuk2020randaugment}, Mixup \citep{zhang2017mixup}, and AugMix \citep{hendrycks2019augmix}. See table \ref{tab:imagenet_hparams} for hyperparameter settings used.

\paragraph{CIFAR-10:} We use CIFAR-10 in sequential and 2D experiments. For sequential, we use the Hyena operator defined in our language tasks and compare with an S4 model \citep{gu2021efficiently} of the same size by swapping layers in the residual blocks. In 2D, we learn Hyena filters (in both $x$ and $y$ dimensions) that are equal to the size of the input shape, and forgo the gating mechanism from our language experiments. We window (i.e., apply a soft mask spatially to) the Hyena filters with a decay term. The rate of decay varies across channels, ensuring different sizes of the filters at initialization. We compare with another implicit 2D convolution, S4ND \citep{nguyen2022s4nd}, by swapping the model layers with the 2D Hyena filters. The "isometric" model consists of 4 residual blocks of model dimension 128. We use basic image augmentations, 0.1 dropout, 0.03 weight decay and train for 100 epochs using a Nvidia T4 GPU.

\begin{table}[ht]
      \small
      \caption{ViT and ViT-Hyena settings for ImageNet-1k).
      }
        \centering
        \begin{tabular}{lcc}
            \toprule
            Image size & $224^2$ \\
            Optimizer & AdamW \\
            Optimizer momentum & $\beta_1,\beta_2=0.9,0.999$ \\
            Weight init & trunc. normal (std=0.02) \\
            ViT base learning rate & $1e^{-3}$ \\
            Hyena-ViT base learning rate & $2e^{-4}$ \\
            ViT weight decay & 0.05 \\
            Hyena-ViT weight decay & 0.01 \\
            Dropout & None \\
            Batch size & 1024 \\
            Training epochs & 300 \\
            Learning rate schedule & cosine decay \\
            Warmup epochs & 10 \\
            Warmup schedule & linear \\
            Randaugment \citep{cubuk2020randaugment} & (9,0.5,layers=2)  \\
            Mixup \citep{zhang2017mixup} & 0.8 \\
            Cutmix \citep{yun2019cutmix} & 1.0 \\
            Random erasing \citep{zhong2020random} & 0.25 \\
            Label smoothing \citep{szegedy2016rethinking} & 0.1 \\
            Stochastic depth \citep{huang2016deep} & 0.1 \\
            Exp.mov. avg (EMA) \citep{polyak1992ema} & None \\
            \bottomrule
        \end{tabular}
        \label{tab:imagenet_hparams}
\end{table}

\section{Theoretical Results and Details}
\subsection{Proofs}
\paragraph{Proof of Proposition \ref{prop:causality}}
\proof A discrete $L$-by-$L$ operator is causal if it is lower triangular, i.e., when there is no leakage of future input information to the output. The ${\sf Hyena}$ operator $\sH$ is the product of alternating diagonal and Toeplitz matrices. Thus, if all the Toeplitz matrices $\sS_h^n$ are lower triangular then $\sH$ is lower triangular. In turn, each $\sS_h^n$ is lower triangular if and only if the filter $h$ is causal, concluding the proof. 
\endproof
\subsection{Analysis of Data-Controlled Mechanisms}\label{app:surrogate_att}
 We discuss the surrogate attention mechanism of {\sf Hyena}-$2$: $q,k,v\mapsto y$:
\begin{equation}\label{eq:linear_attention}
    \begin{aligned}
        z_t &= k_t(\varphi * v)_t \\
        y_t &= q_t(\psi * z)_t
    \end{aligned}
\end{equation}
If $\varphi$ and $\psi$ are convolutions parametrized via state-space models (SSMs), the above resembles the H3 mechanism \citep{dao2022hungry}. We investigate the effect of the convolutional kernels $\varphi$ and $\psi$ on the attention layer. We start by introducing a matrix representation of the layer, and we isolate the \textit{attention matrix} $\sA_\varphi^\psi(q,k)$ such that
\begin{equation}\label{eq:linear_attention_matrix}
    \begin{aligned}
        y &= \sA_\varphi^\psi(q,k)v.
    \end{aligned}
\end{equation}
\paragraph{Isolating the surrogate attention matrix}
In the case of length-$L$ discrete sequences
\begin{equation}\label{eq:linear_attention_convolution3}
    \begin{aligned}
        z_t &= k_t \sum_{m=0}^{L-1} \varphi_{t-m} v_m \\
        y_t &= q_t \sum_{m=0}^{L-1} \psi_{t-m} z_m\\
    \end{aligned}
\end{equation}
Therefore we can rewrite \eqref{eq:linear_attention} as
\begin{equation}
    \begin{aligned}
        y_t &= q_t \sum_{m=0}^{L-1} \psi_{t-m} k_m \sum_{n=0}^{L-1} \varphi_{m-n} v_n\\
             &= q_t \sum_{m=0}^{L-1} \sum_{n=0}^{L-1} \psi_{t-m} k_m \varphi_{m-n} v_n &&\quad\text{Move $\psi$, $k$ inside inner sum}\\
             &= q_t \sum_{n=0}^{L-1} \sum_{m=0}^{L-1} \psi_{t-m} k_m \varphi_{m-n} v_n &&\quad\text{Index shift}\\
             &= \sum_{n=0}^{L-1} q_t \sum_{m=0}^{L-1} \psi_{t-m} k_{m} \varphi_{m-n} v_{n}\\
    \end{aligned}
\end{equation}
And we can define the surrogate attention matrix $\sA_\varphi^\psi(q,k)$
\begin{equation}\label{eq:linear_attention_matrix2}
    \begin{aligned}
        [\sA_\varphi^\psi(q,k))]_{t,t'} &= q_t \sum_{m=0}^{L-1} \psi_{t-m} k_{m} \varphi_{m - t'}. \\
    \end{aligned}
\end{equation}
\begin{tcolorbox}[enhanced, drop fuzzy shadow, breakable, frame hidden, sharp corners] {\bf Continuous Signals:} 
    We can also consider the case of continuous signals on a group $G$. In the continuous case, we can expand the convolutions in \eqref{eq:linear_attention} as
    \begin{equation}\label{eq:linear_attention_convolution}
        \begin{aligned}
            (\varphi * v)_t = \int_G \varphi_{t-g} v_g \dd g,\qquad
            (\psi * z)_t = \int_G \psi_{t - g} z_g \dd g
        \end{aligned}
    \end{equation}
    This allows us to rewrite \eqref{eq:linear_attention} as
    \begin{equation}\label{eq:linear_attention_convolution2}
        \begin{aligned}
            y_t &= q_t(\psi * k(\varphi * v))_t \\
            &= q_t \int_G \psi_{t-g} \left[ k_g\int_G \varphi_{g - \tau} v_\tau \dd \tau \right] \dd g \\
            &= q_t \int_G \left[ \int_G \psi_{t-g} k_g \varphi_{g - \tau} v_\tau \dd \tau \right] \dd g\\
            &= q_t \int_G \left[ \int_G \psi_{t-g} k_g \varphi_{g - \tau} v_\tau \dd g \right] \dd \tau &&\quad\text{Variable swap}\\
            &= \int_G \left[ q_t \int_G \psi_{t-g} k_g \varphi_{g - \tau} v_\tau \dd g \right] \dd \tau &&\quad\text{Pull $q_t$ in $\tau $ integral}\\
            & = \int_G \left[ q_t \int_G \psi_{t-g} k_g \varphi_{g - \tau} \dd g \right] v_\tau \dd \tau &&\quad\text{Pull $v_\tau$ out of $g$ integral}.
        \end{aligned}
    \end{equation}
    There is a linear operator $\cA: v \mapsto y=\cA v$ which we interpret as the surrogate attention operator. $\cA$ is conditioned on the \textit{query} $q$, \textit{key} $k$ and filters $\varphi$ and $\psi$, $\cA = \cA_{\varphi}^\psi(q,k)$. The kernel $\mathcal K$ of the operator is given by
    \begin{equation}
        \begin{aligned}
            \mathcal K(t,t') &= q_t\int_G \psi_{t-g} k_g \varphi_{g - t'} \dd g \\
        \end{aligned}
    \end{equation}
    %
        %
\end{tcolorbox}
\paragraph{Operator decomposition of the surrogate attention matrix}
We can decompose the linear map $v\mapsto y;~ y = \sA_\varphi^\psi(q,k)v$ into a sequence of factors, each dependent on a projection of the input $\sA_\varphi^\psi(q,k) = \sA^\psi(q) \sA_\varphi(k)$. Let $\sD_q$ and $\sD_k$ be the $L$-by-$L$ diagonal matrices whose respective main diagonal entries are the respective entries of $q$ and $k$. Then, we have that
\begin{equation}\label{eq:linear_attention_matrix3}
    \begin{aligned}
        \sA^\psi(q) &= \sD_q \sS_\psi,\qquad \sD_q = \diag(q), \\
        \sA_\varphi(k) &= \sD_k\sS_\varphi,\qquad \sD_k = \diag(k).
    \end{aligned}
\end{equation}
The matrix has been decomposed into two terms $\sA^\psi(q)$ and $\sA_\varphi(k)$ constructed by multiplying the diagonal matrices $\sD_q$ and $\sD_k$ with the Toeplitz matrices $\sS_\psi$ and $\sS_\varphi$. $\sS_\psi$ and $\sS_\varphi$ are the kernels of the convolution operators with filter's impulse responses $\psi$ and $\varphi$ respectively. In the current applications of interest, $\psi$ and $\varphi$ are chosen to be causal, i.e. $\psi[t]=0 \text{ for } t<0$ and $\varphi[t]=0 \text{ for } t<0$. This results in $\sS_\psi$ and $\sS_\varphi$ to be lower triangular matrices 
\begin{equation}
    \sS_\psi = \begin{bmatrix}
        \psi_0 & 0 & \cdots & 0 \\
        \psi_1 & \psi_0 & \cdots & 0 \\
        \vdots & \ddots & \ddots & \vdots \\
        \psi_{L-1} & \psi_{L-2} & \cdots & \psi_0
    \end{bmatrix}, \qquad
    \sS_\varphi = \begin{bmatrix}
        \varphi_0 & 0 & \cdots & 0 \\
        \varphi_1 & \varphi_0 & \cdots & 0 \\
        \vdots & \ddots & \ddots & \vdots \\
        \varphi_{L-1} & \varphi_{L-2} & \cdots & \varphi_0
    \end{bmatrix}.
\end{equation}
The surrogate attention matrix is then given by
\begin{equation}
    \sA_\varphi^\psi(q,k) = \sD_q \sS_\psi \sD_k \sS_\varphi
\end{equation}
We can expand the matrix multiplications in \eqref{eq:linear_attention_matrix3} in the case of causal filters $\varphi$ and $\psi$ as
\begin{equation}\label{eq:linear_attention_matrix4}
    \begin{aligned}
        \underset{ 
        \begin{bmatrix}
            q_0 &  &  &  \\
             & q_1 &  &  \\
             &  & \ddots &\\
             &  &  & q_{L-1}
        \end{bmatrix}}{\displaystyle \sD_q}
        \underset{ 
        \begin{bmatrix}
            \psi_0 & & &  \\
            \psi_1 & \psi_0 & &  \\
            \vdots & \ddots & \ddots & \\
            \psi_{L-1} & \psi_{L-2} & \cdots & \psi_0
        \end{bmatrix}}{\displaystyle \sS_\psi}
        \underset{ 
        \begin{bmatrix}
            k_0 &  &  &  \\
             & k_1 &  &  \\
             &  & \ddots &\\
             &  &  & k_{L-1}
        \end{bmatrix}}{\displaystyle \sD_k}
        \underset{
        \begin{bmatrix}
            \varphi_0 & & &  \\
            \varphi_1 & \varphi_0 & &  \\
            \vdots & \ddots & \ddots & \\
            \varphi_{L-1} & \varphi_{L-2} & \cdots & \varphi_0
        \end{bmatrix}}{\displaystyle \sS_\varphi} 
        &\\
        \underset{\displaystyle \sA_\psi(q)}{ 
        = \begin{bmatrix}
            q_0 \psi_0 &  &  &  \\
            q_1 \psi_1 & q_1 \psi_0 &  &  \\
            \vdots & \ddots & \ddots & \\
            q_{L-1} \psi_{L-1} & q_{L-1} \psi_{L-2} & \cdots & q_{L-1} \psi_0
        \end{bmatrix}}
        \underset{\displaystyle \sA_\varphi(k)}{ 
        \begin{bmatrix}
            k_0 \varphi_0 &  &  &  \\
            k_1 \varphi_1 & k_1 \varphi_0 &  &  \\
            \vdots & \ddots & \ddots & \\
            k_{L-1} \varphi_{L-1} & k_{L-1} \varphi_{L-2} & \cdots & k_{L-1} \varphi_0
        \end{bmatrix}} &\\
    \end{aligned}
\end{equation}
\begin{tcolorbox}[enhanced, breakable, frame hidden, drop fuzzy shadow, sharp corners] {\bf Fourier decomposition of convolution operators:} The kernels of the convolution operators $\sS_\psi$ and $\sS_\varphi$ are diagonalized by the Fourier transform matrix $\sW\in\bC^{L\times L},~ \sW_{nm} = z^{m}, ~ z = e^{j 2\pi n / L}$. The Fourier transform of the convolution operator $\sS_\psi$ is given by
\begin{equation}
    \sS_\psi = \sW^* \sD_\Psi \sW, \quad \sS_\Phi = \sW^* \sD_\Phi \sW
\end{equation}
where $\sD_\Psi, \sD_\Phi\in\bC^{L\times L}$ are diagonal matrices constructed from the frequency responses (the \textit{discrete Fourier transform}) $\Psi=\sW\psi,\Phi=\sW\varphi$, respectively. This decomposition can be used to simplify the matrix multiplication in \eqref{eq:linear_attention_matrix4}:
\begin{equation}
    \sA = \sD_q \sS_\psi \sD_k \sS_\varphi = \sD_q  \sW^* \sD_\Psi \sW \sD_k \sW^* \sD_\Phi \sW
\end{equation}
An important property of the above is the non-commutativity of $\sD_q$ and $\sS_k$ with $\sW*$. If the two operators commuted, we would obtain
\begin{equation}
    \boxed{
    \sA =  \sD_q  \sW^* \sD_\Psi \sW \sD_k \sW^* \sD_\Phi \sW = \sW^* \sD_q \sD_\Psi \sD_k \sD_\Phi \sW
    }
\end{equation}
which reduces the entire layer to a simple convolution. The non-commutativity of the \textit{gating} term acts as a non-linearity in chain of convolution operators.
\end{tcolorbox}

\section{Discussion and Additional Results}\label{app:add_results}

\paragraph{Vocabulary size scaling}

Table \ref{scaling_vsize} showcases interesting correlation between associative recall performance for varying vocabulary sizes and loss on the {\sc The Pile}. In this case, we fix sequence length for associative recall to be $2048$, the same sequence length used to train all models on the {\sc The Pile}.

We observe a similar phenomenon on other slices of tasks from our mechanistic design benchmarks, indicating that it may be possible to derive predictive laws for performance at scale, based on fast experimentation on synthetic tasks with models of $1$ or $2$ layers. Surprisingly, performance on our language synthetics appears to be further linked to performance as attention replacement in other domains (Appendix \ref{appendix:image-classification} for results on image classification).

\begin{table}[!bh]
\small
\centering
\caption{{\sf Hyena} Accuracy on associative recall with varying vocabulary size $10$, $20$, $30$, $40$ in relation to test loss on {\sc The Pile} after $5$ billion tokens. We notice a correlation between the two performance metrics, suggesting that slices of our mechanistic design synthetics may be potentially predictive of performance at scale.}
\vspace{2mm}
\label{scaling_vsize}
\setlength{\tabcolsep}{4pt}
\begin{tabular}{@{}c|ccccc@{}}
\toprule
Model & Acc @ $10$ & Acc @ $20$ & Acc @ $30$ & Acc @ $40$ & Loss @ $5$B on {\sc The Pile} \\
\midrule 
Conv1d & $32$ & $11$ & $10$ & $8$ & $4.21$\\ 
AFT-conv & $55$ & $21$ & $12$ & $10$ & $3.57$\\ 
H3 & $92$ & $60$ & $13$ & $10$ & $2.69$\\
Transformer & $100$ & $100$ & $92$ & $82$ & $2.59$\\ 
{\sf Hyena} & $100$ & $100$ & $98$ & $85$ & $2.59$\\ 
\bottomrule
\end{tabular}
\end{table}

\paragraph{Single layer recall}

All experiments on our synthetic tasks default to $2$ layer models. We choose $2$ as it is the canonical number for mechanistic analysis of Transformers \citep{elhage2021mathematical} based on \textit{circuits}. Interestingly, a single layer of {\sf Hyena} (width $64$) is capable of performing associative recall, solving the task completely even in the challenging setting with vocabulary size $40$. Reverse engineering exactly how the single {\sf Hyena} operator is able to perform recall is left for future work.

\subsection{Learning Arithmetic}
We showcase an additional task in our mechanistic design benchmark: learning arithmetic. We train {\sf Hyena} models of increasing depth ($1$, $2$ and $3$ layers) on a dataset of $D_n$-digit addition. As an example, a $3$-digit addition input sample is given by the sequence
\[ 
    {\tt 1, 2, 3, 9, 5, 4, 1, 0, 7, 7}
\]
where the first $6$ digits contain the two $3$ digits numbers to add, and the last $4$ the result. Our models are optimized using standard autoregressive training i.e., predicting the next token, since they are causal. In particular, we optimize models to learn a map $x \mapsto y$ where $x$ is the original prompt without the last element, and $y$ equal to $x$ shifted right by one position. We mask the first $2 D_n - 1$ elements of the loss for each sequence since they contain predictions for addends and not results.

\begin{figure}
    \centering
    \input{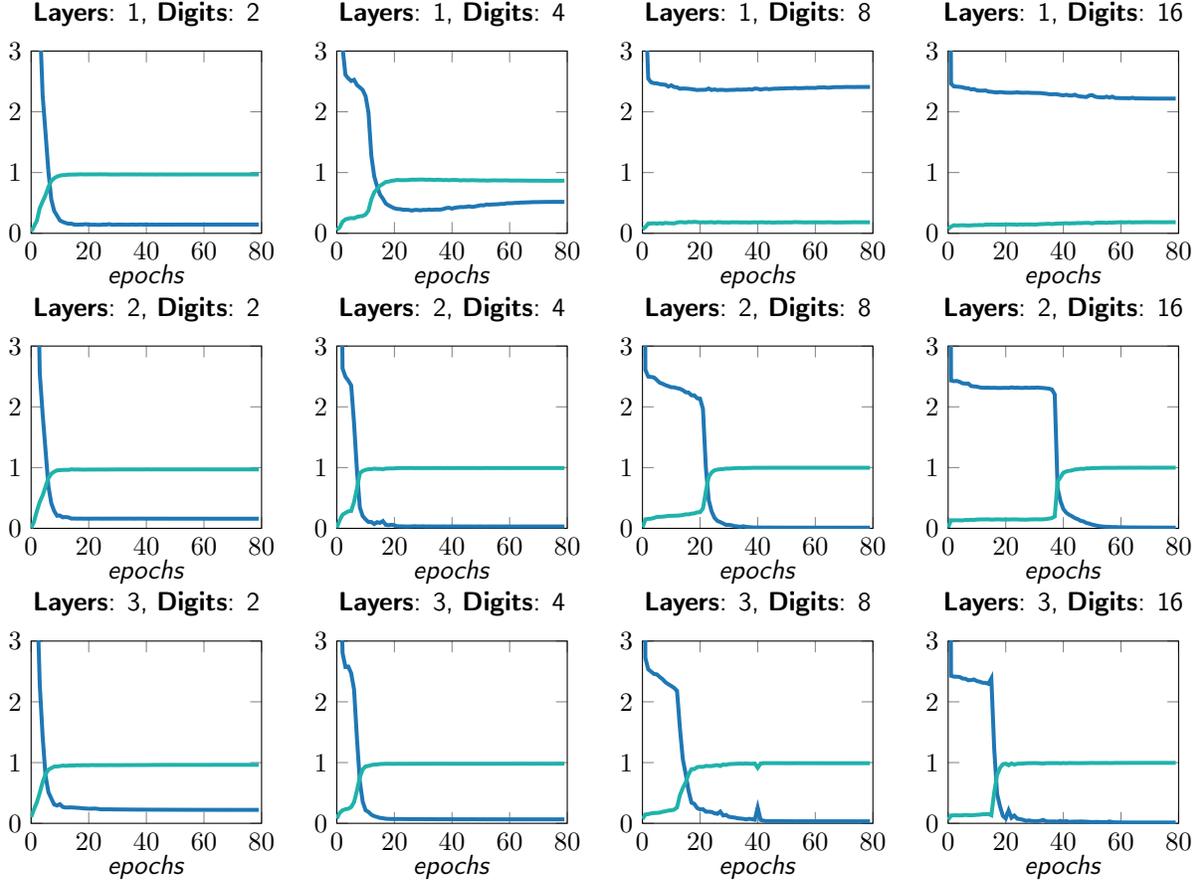}
    \vspace{-2mm}
    \caption{Test loss and accuracy of $\sf Hyena$ on addition with different numbers of digits and model depths. Each plot reports the results of a different experiment, with the curve tracing test results during training.}
    \label{fig:arithmetic}
\end{figure}

We report results in Figure \ref{fig:arithmetic}. A single layer of {\sf Hyena} is able to learn to perform addition with up to $4$ digits. Longer numbers require deeper models. In our experiments, alternative architectures such as AFT-conv struggle to learn arithmetic, signaling a cap in capability.

\section{Samples and Visualizations}

\subsection{Hyena Matrices}
\begin{figure}
    \centering
    \includegraphics[width=\linewidth]{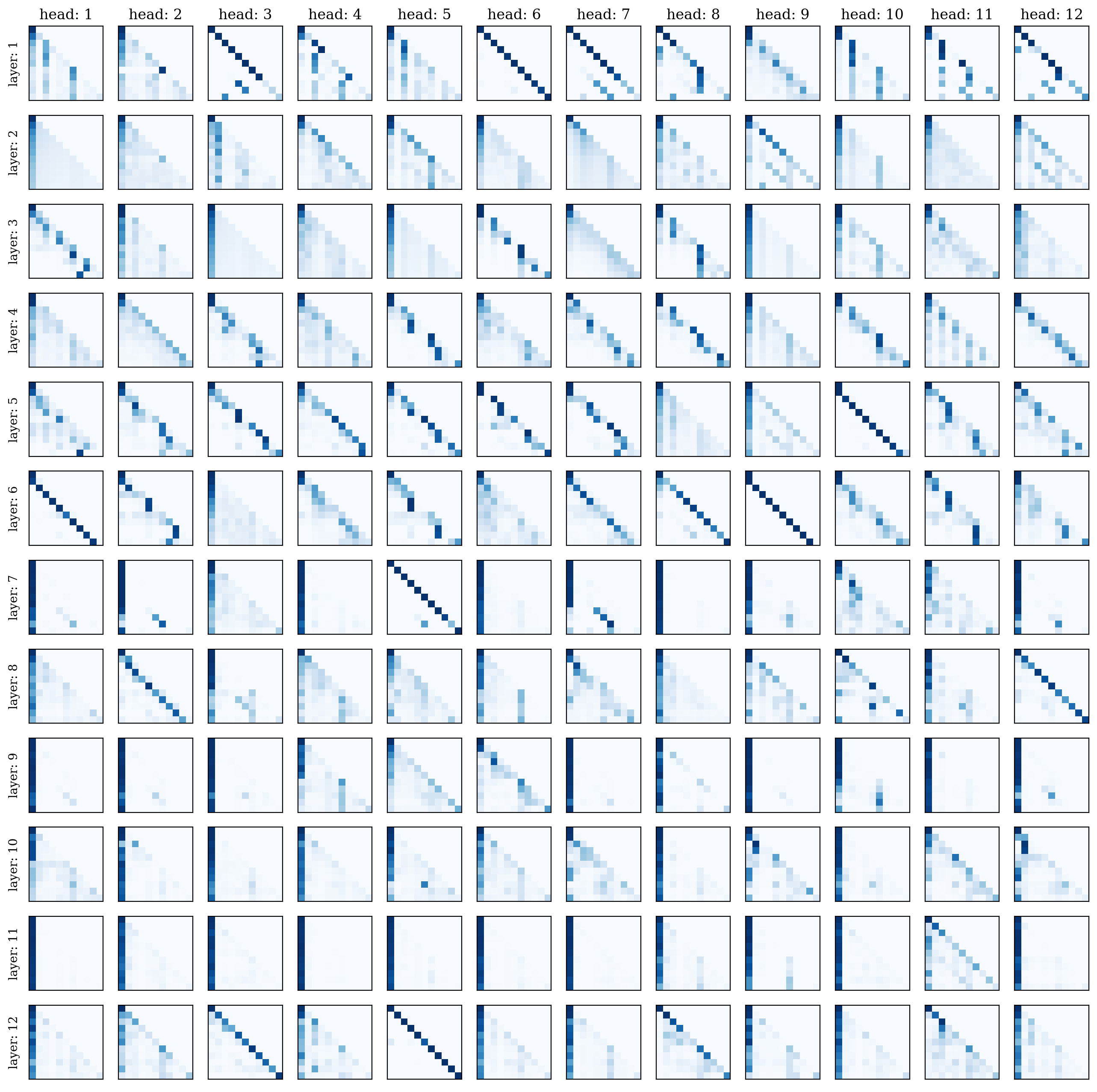}
    \vspace{2mm}
    \caption{Attention matrices from a GPTNeo small model. "We use the test string "\textit{Attention is all you need. Attention is}".}
    \label{fig:comparisons_1}
\end{figure}

\begin{figure}
    \centering
    \includegraphics[width=\linewidth]{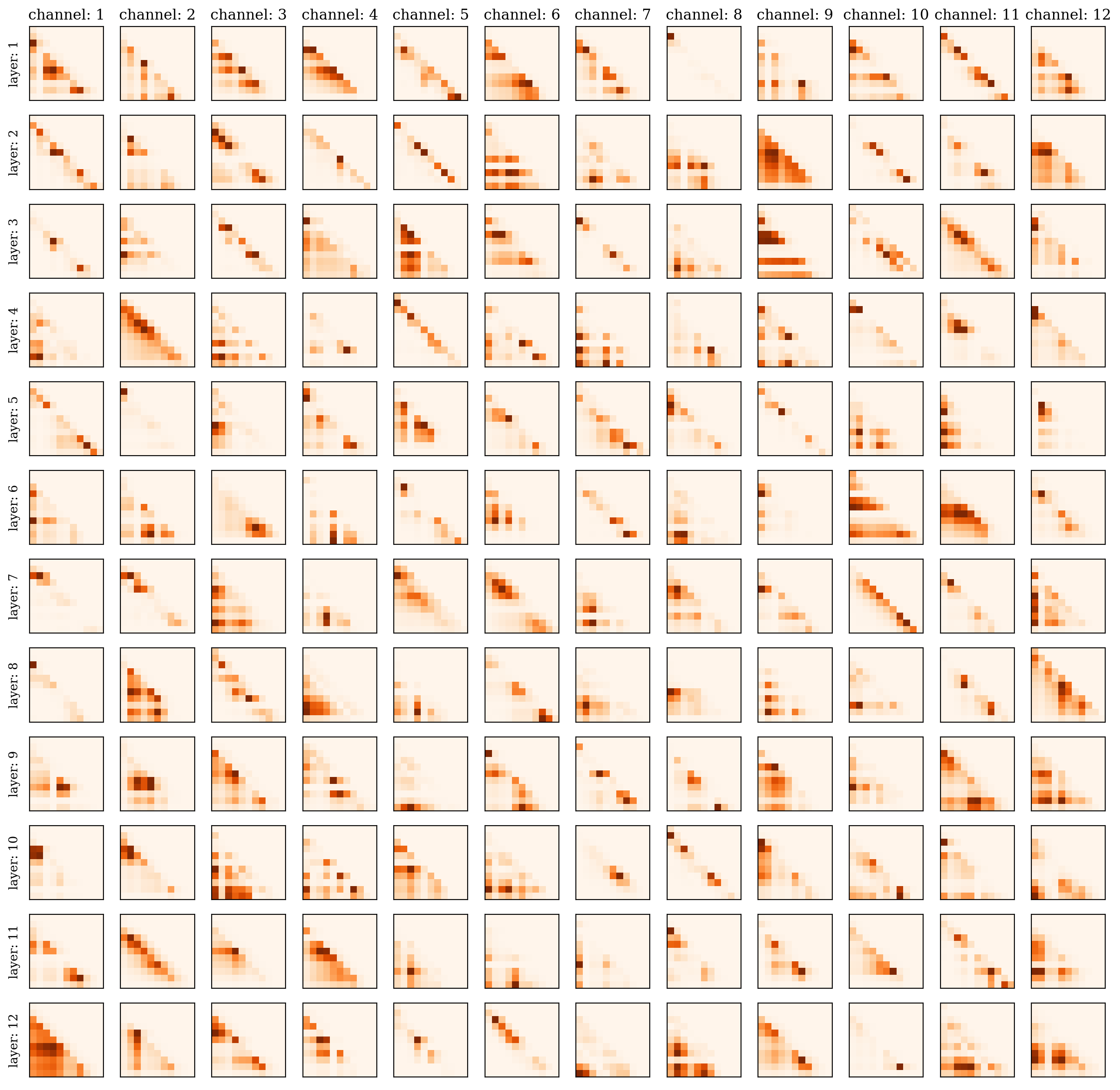}
    \caption{Hyena matrices from a {\sf Hyena} small (same model used for SuperGLUE downstream evaluations). "We use the test string "\textit{Attention is all you need. Attention is}". We note that {\sf Hyena} has a different data-controlled matrix for each \textit{channel} i.e. for each dimension in its width, since it does not use heads.}
    \label{fig:comparisons_2}
\end{figure}

\begin{figure}[h]
    \centering
    \includegraphics[width=\linewidth]{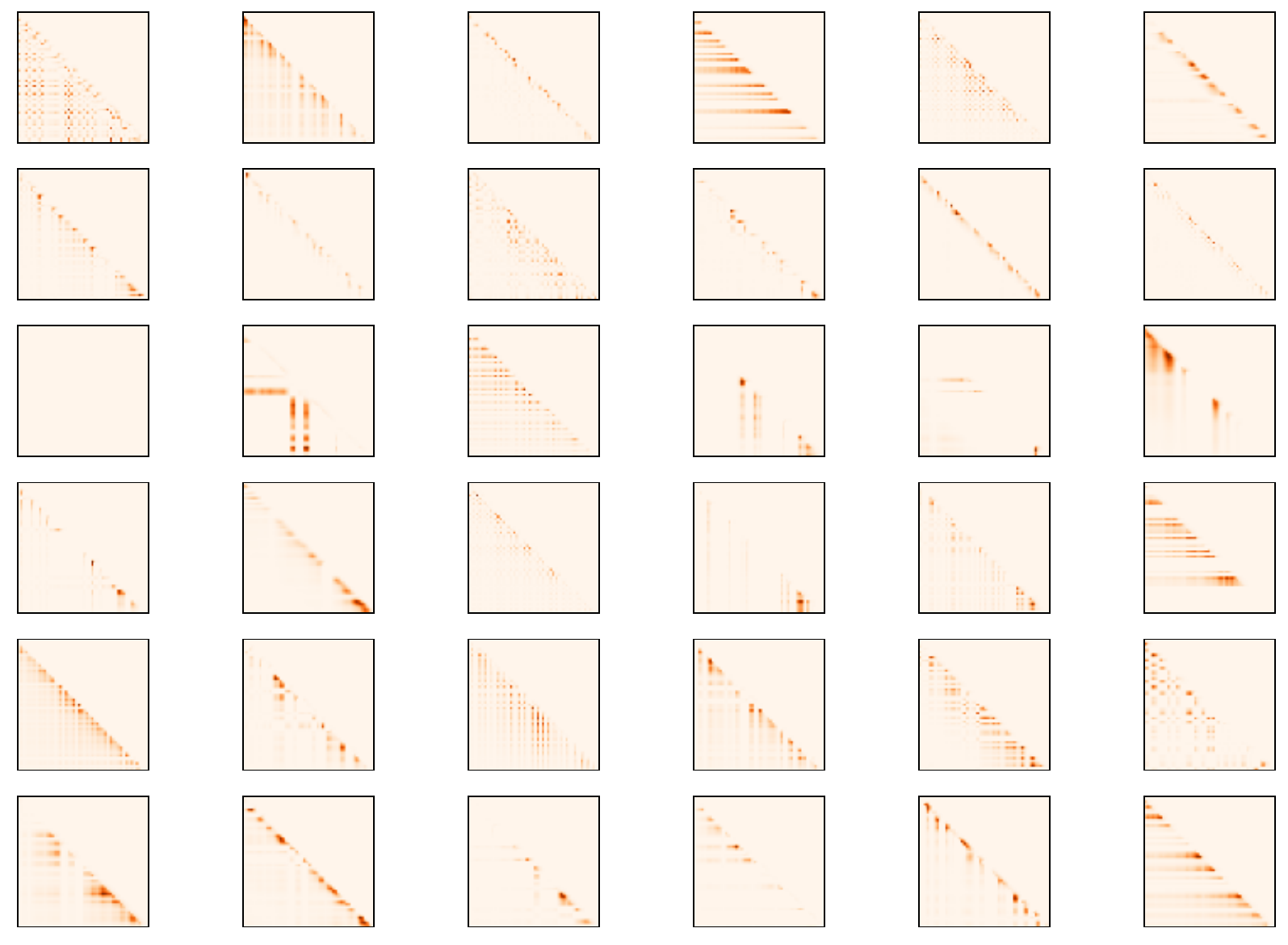}
    \caption{Data-controlled ${\sf Hyena}$ matrices ($355$M model), activated by the string "\textit{When a doctor doctors a doctor, does the doctor doing the doctoring doctor as the doctor being doctored wants to be doctored or does the doctor doing the doctoring doctor as they want to doctor?}". Rows in the plot are matrices from different layers, columns are matrices from different channels. The operator shows characteristic patterns of attention matrices, without attention.}
    \label{fig:visu}
\end{figure}

\begin{figure}[h]
    \centering
    \includegraphics[width=\linewidth]{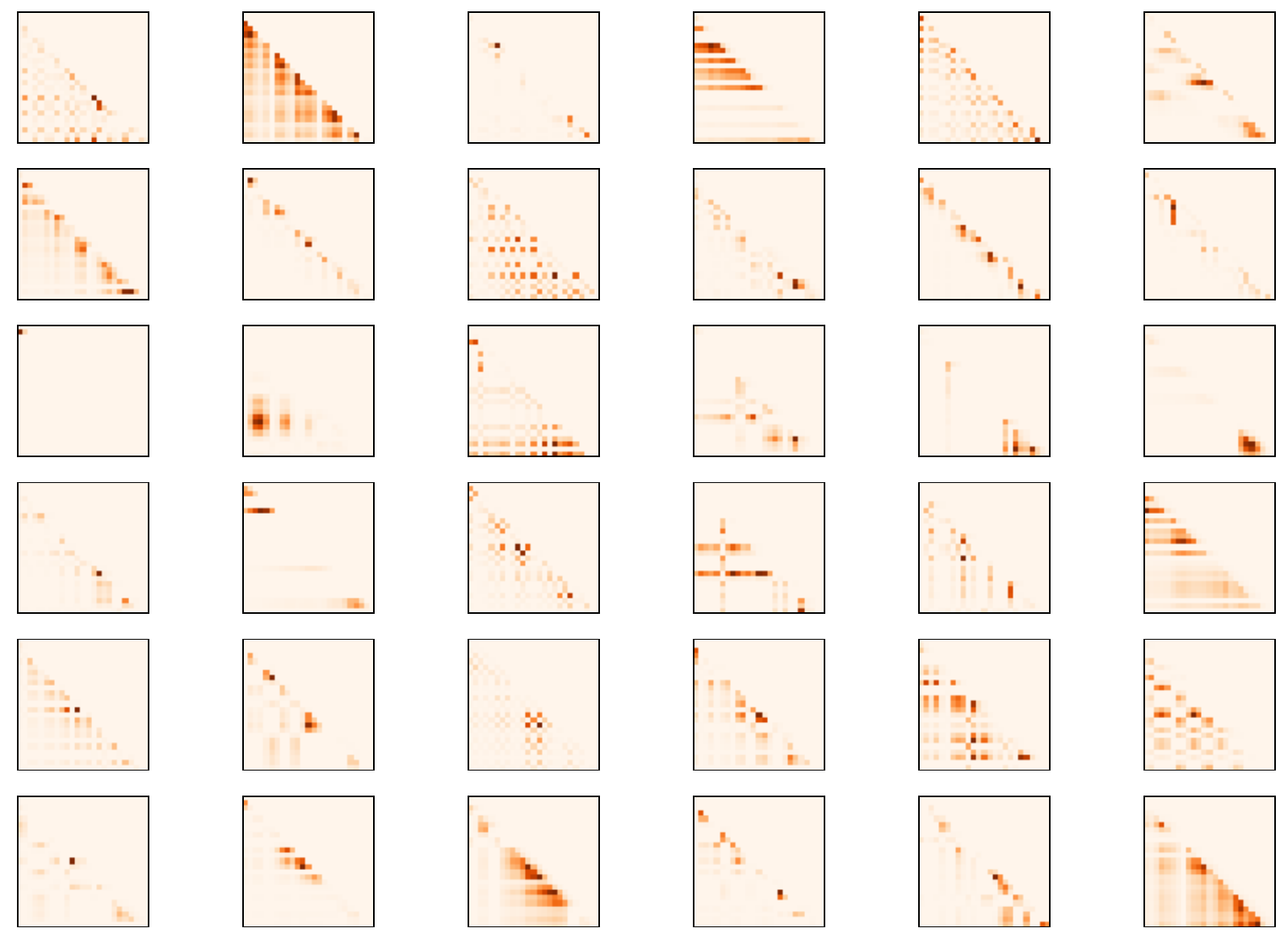}
    \caption{Data-controlled ${\sf Hyena}$ matrices ($355$M model), activated by the string "\textit{Mrs. Dursley, Mr. Dursley, Dudley Dursley}", from \href{ttps://www.lesswrong.com/posts/j6s9H9SHrEhEfuJnq/causal-scrubbing-results-on-induction-heads}{\textit{Causal scrubbing: results on induction heads}}. Rows in the plot are matrices from different layers, columns are matrices from different channels.}
    \label{fig:visu2}
\end{figure}

We provide visualizations of attention and ${\sf Hyena}$ matrices activated by test strings. In~\ref{fig:comparisons_1},~\ref{fig:comparisons_2}, we compare GPTNeo \citep{gpt-neo} attention matrices with Hyena matrices extracted by our pre-trained small {\sf Hyena} model. In \ref{fig:visu} and \ref{fig:visu2}, we provide additional Hyena matrices for the $355$M model, activated by test strings of different length.

For attention, we visualize the raw post-softmax matrix. For ${\sf Hyena}$ matrices, we plot the (element-wise) absolute value of $\sH(u)$:
\[ 
\begin{aligned}
\sH(u) &= \sD_x^N\sS_h^N \cdots \sD_x^2\sS_h^2\sD_{x}^1\sS_h^1 \\ 
\hat \sH(u)_{ij} &= \left|\sH(u)_{ij}\right|
\end{aligned}
\]
Since {\sf Hyena} does not normalize the entries of its matrices with e.g., softmax, there are notable differences with attention: (1) the entries of $\sH(u)$ can be either positive and negative, and (2) the magnitude is unconstrained. We observe the magnitude of matrices in pre-trained {\sf Hyena} models to be around $10^{-3}$.



\clearpage 

\subsection{Hyena Filters}

\begin{figure}
    \centering
    \includegraphics[width=0.99\linewidth]{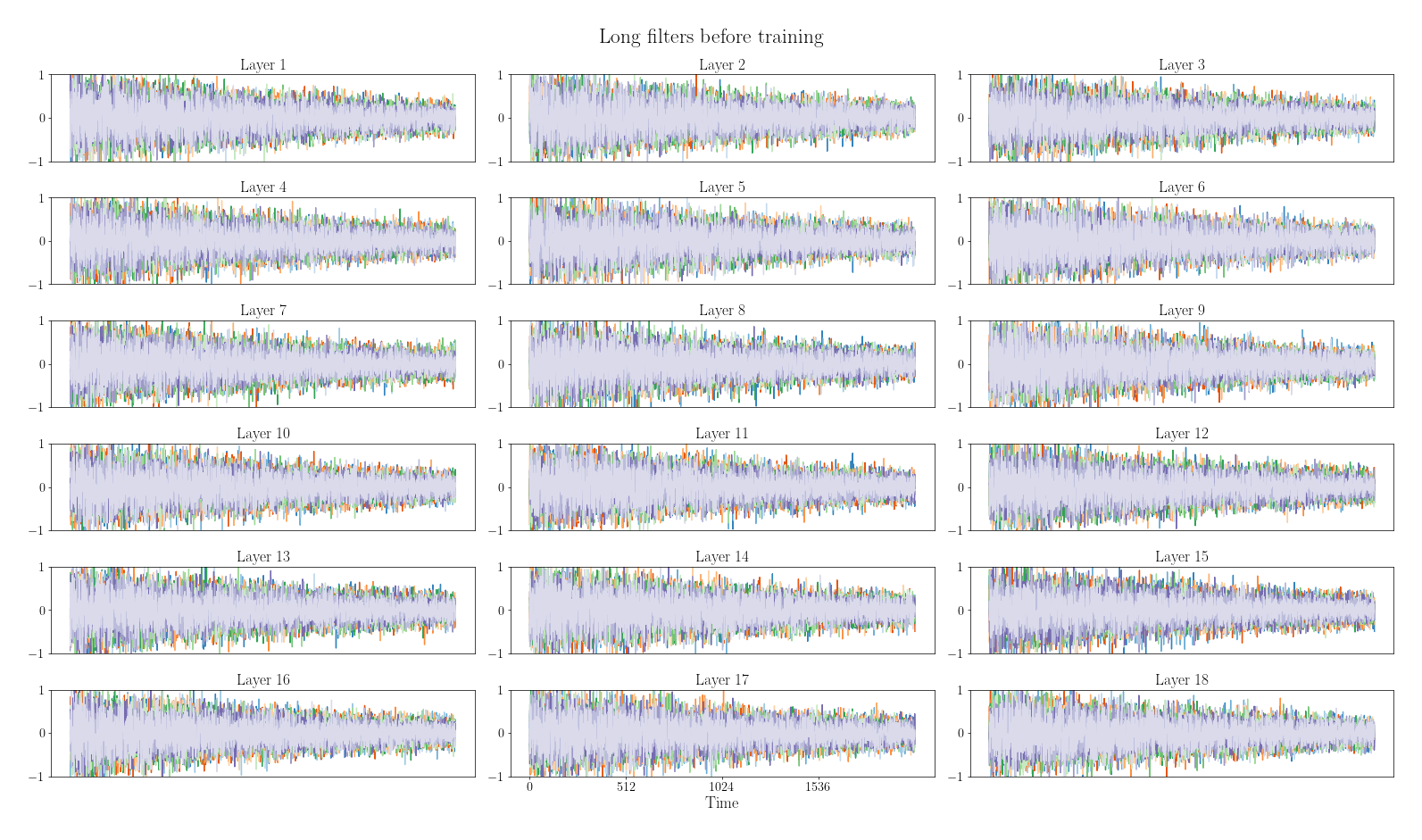}
    \vspace{2mm}
    \includegraphics[width=0.99\linewidth]{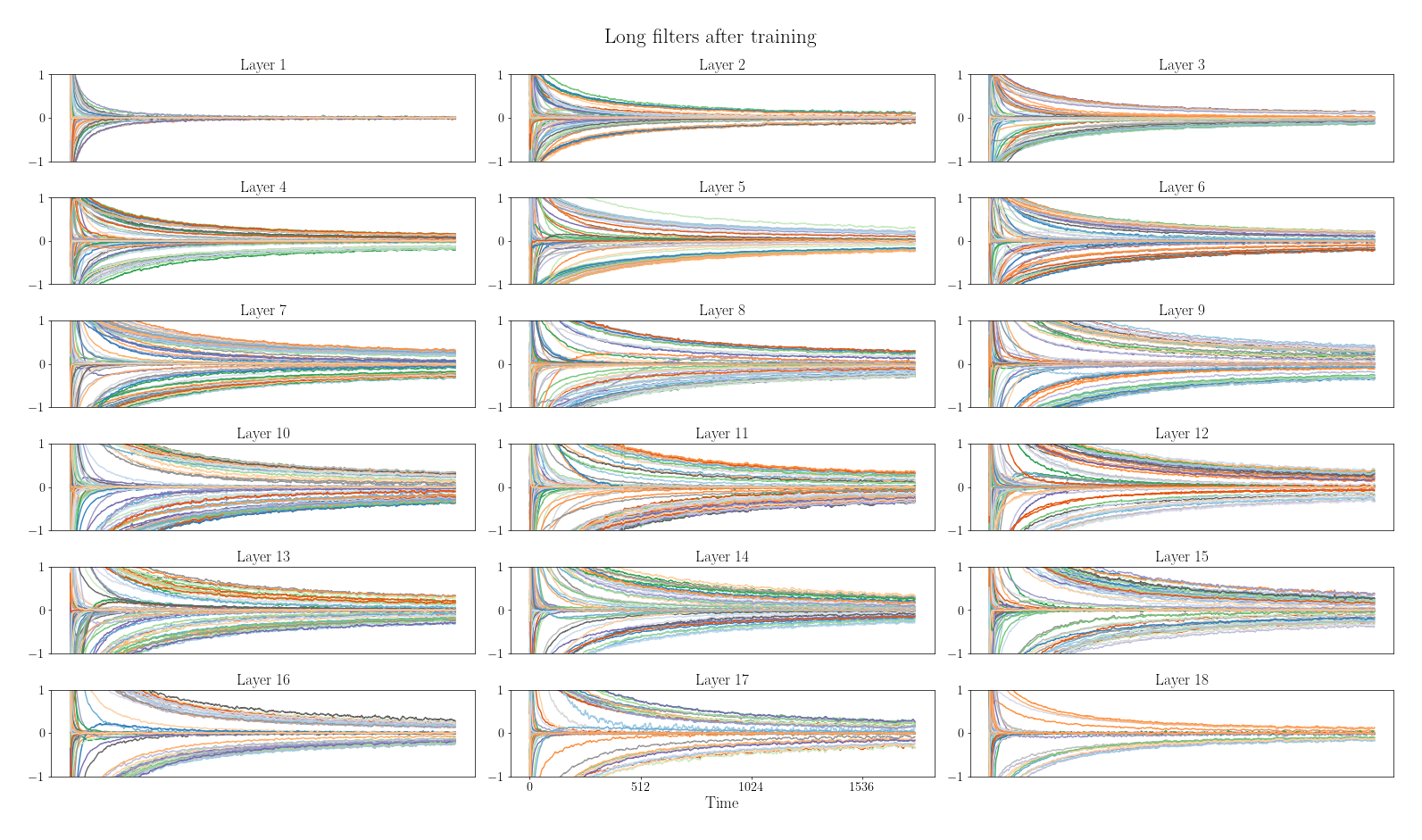}
    \caption{\textbf{[Top]}: Long convolution {\sf Hyena} filters at initialization ($153$M parameters, $18$ layer model). \textbf{[Bottom]}: Filters after training for $130$ billion tokens on {\sc The Pile}.}
    \label{fig:hyena_filters}
\end{figure}

Figure \ref{fig:hyena_filters} provides a visualization of {\sf Hyena} long convolution filters at initialization and after training to completion on {\sc The Pile}. 

 We find a substantial performance difference (up to $5\%$ perplexity) between initialization schemes. If the filters at initialization are excessively smooth (see Appendix \ref{app:posemb} for a discussion of positional encoding and activation), the model finds a worse solution and takes longer to converge. Further, we observe initialization schemes that regularize filters towards typical filters learned at convergence to decrease performance. These observations are in line with performance gaps between convolution parametrization schemes discussed in main text and Appendix \ref{app:icl}. In particular, the performance improvements obtained via {\sf Hyena} filters could be due to easier optimization in the space of convolutional filters.

 At convergence, {\sf Hyena} learns a collection of lower-order filters with a similar structure, which can be exploited to further speed up inference after training.

\subsection{Positional Encoding and Filters Initialization}\label{app:posemb}
\begin{figure}
    \centering
    \includegraphics[width=.8\linewidth]{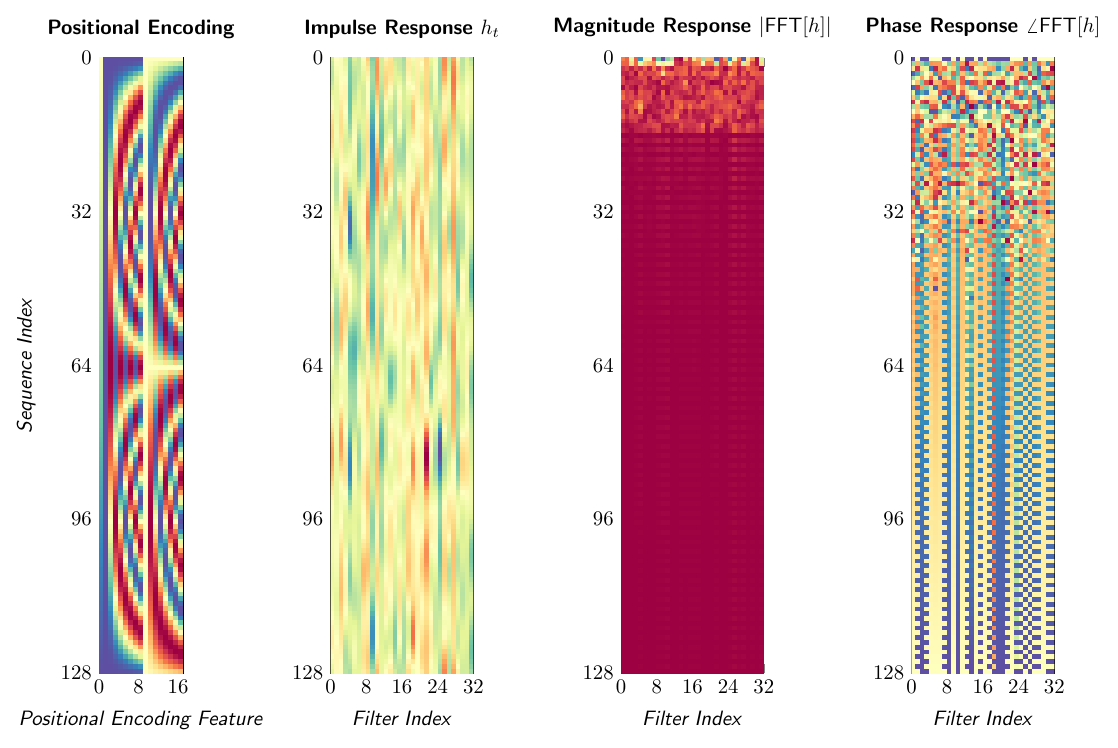}
    \caption{$\sf Hyena$ filters at initialization with 17 positional encoding features $K=8$.}
    \label{fig:pos_enc_17_1}
\end{figure}

\begin{figure}
    \centering
    \includegraphics[width=.8\linewidth]{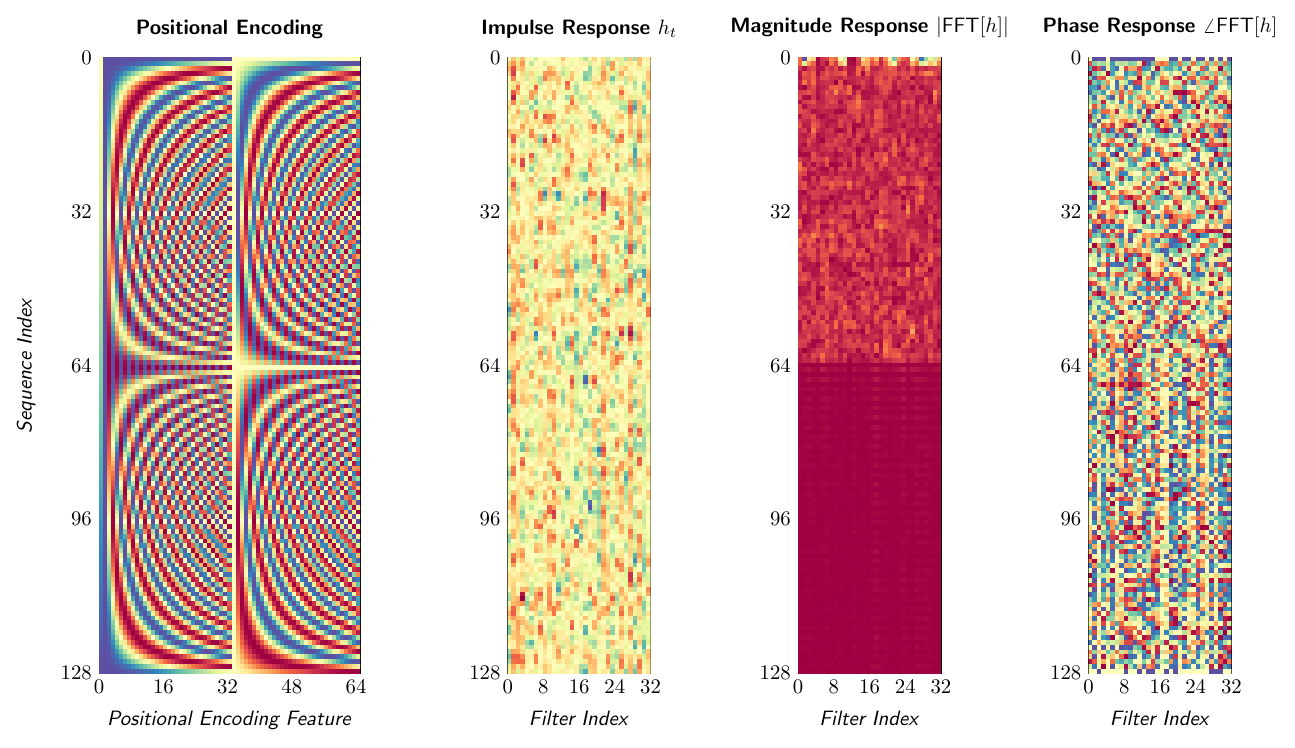}
    \caption{$\sf Hyena$ filters at initialization with 65 positional encoding features $K=32$.}
    \label{fig:pos_enc_65_1}
\end{figure}

\begin{figure}
    \centering
    \includegraphics[width=.9\linewidth]{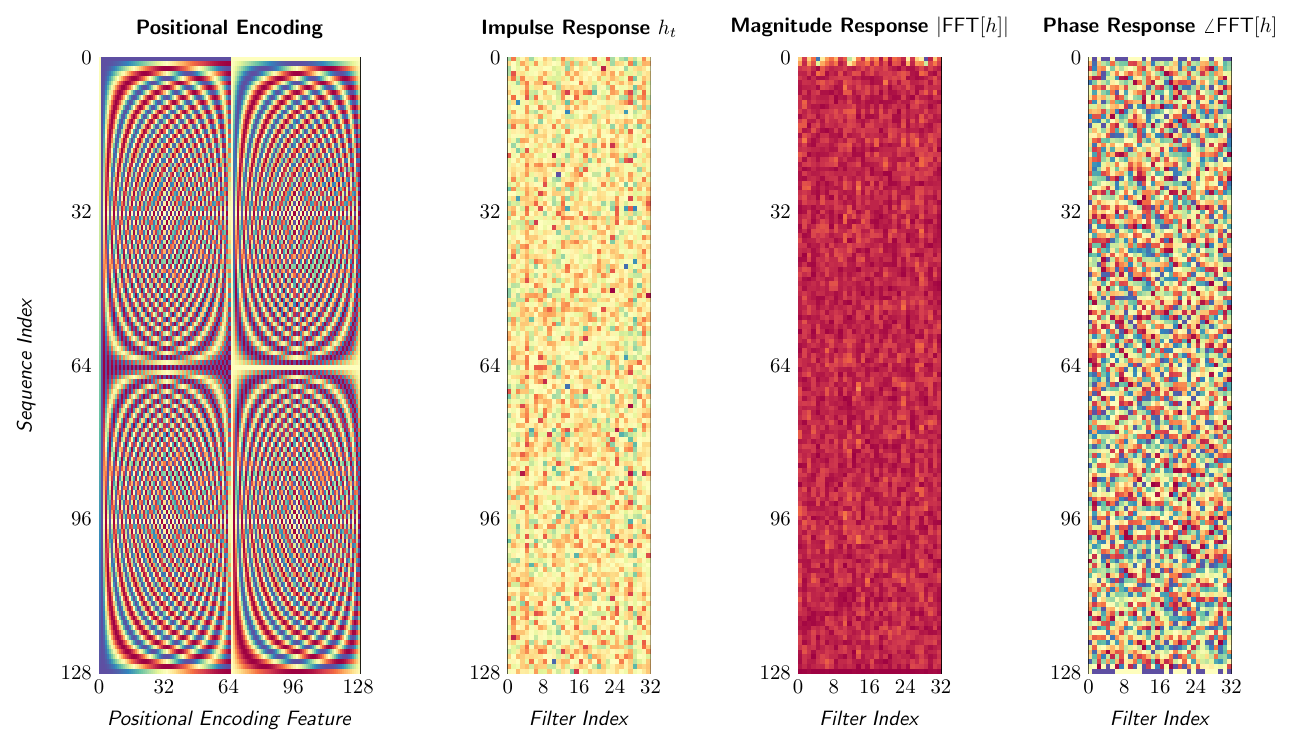}
    \caption{$\sf Hyena$ filters at initialization with 65 positional encoding features $K=64$.}
    \label{fig:pos_enc_128_1}
\end{figure}

The positional encoding chosen for the $\sf Hyena$ filters is a truncated complex exponential basis. Specifically, with $\rho_k(t) = e^{i2\pi kt/L}$ for $k=0,\dots K-1$, the positional encoding is defined as a map from $\R$ to $\R^{2K+1}$ such that
\[
    {\sf PositionalEncoding}(t) = 
    \begin{bmatrix}
        t&
        \mathfrak{R}[\rho_0](t)&
        \cdots&
        \mathfrak{R}[\rho_{K-1}](t)&
        \mathfrak{I}[\rho_0](t)&
        \cdots&
        \mathfrak{I}[\rho_{K-1}](t)
    \end{bmatrix}
\]
where $\mathfrak{R}[\cdot]$, $\mathfrak{I}[\cdot]$ denote the real and imaginary part of their argument, respectively. In the main text, we use $D_e = 2K + 1$ to denote the size of a positional encoding with $K$ features. The number of features of the positional encoding has an impact on the filter initialization and training performances. In particular, we show how $K$ leads to a preconditioning of the spectrum of the filter at initialization. Figures~\ref{fig:pos_enc_17_1},~\ref{fig:pos_enc_65_1},~\ref{fig:pos_enc_128_1} display the initialized filters (with no $\sf Window$ function) for different values of $K$ ($\{8, 32, 64\}$) for $L=128$ and frequency $\omega_a$ of sinusoidal activation $\sigma(\cdot) = \sin(\omega_a \cdot)$ set to 1. We notice how the choice of $K$ induces a bias in the modeled frequencies at initialization. Specifically the filters resemble low-pass filters with a cut-off frequency of approximatively $2K + 1$. 

\begin{figure}
    \centering
    \includegraphics[width=.9\linewidth]{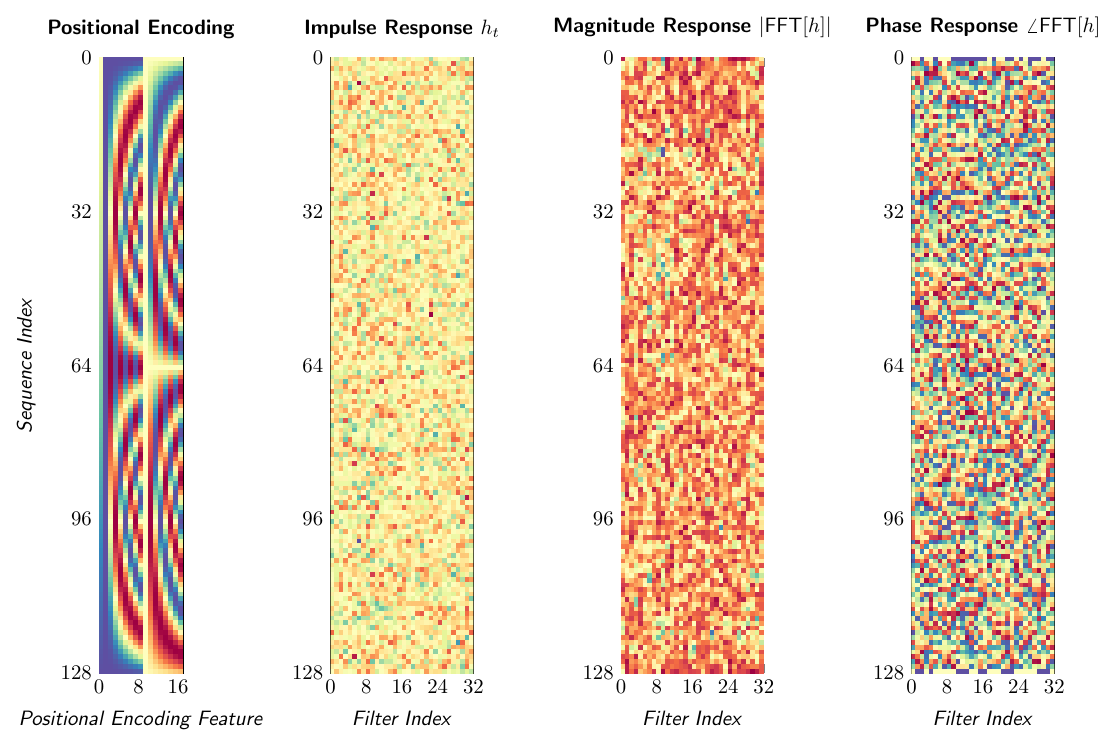}
    \caption{$\sf Hyena$ filters at initialization with 17 positional encoding features $K=8$ and frequency of sinusodial activation set to 10.}
    \label{fig:pos_enc_17_10}
\end{figure}

This cut-off frequency is strongly related to the \textit{smoothness} of the filter; as previously mentioned, we empirically observe better training dynamics of filters initialized to be non-smooth, i.e. with a rich high-frequency content. While we can achieve good initializations by increasing $K$, this results in larger $\sf FFN$s (its input dimension is $2K + 1$, i.e. the number of positional encoding features) which come with a higher parameter count. A more efficient solution is to increase the frequency $\omega_a$ of the sinusoidal activation. Figure~\ref{fig:pos_enc_17_10} show how with $K=8$ we can cover the full spectrum simply by setting $\omega_a=10$. 


\clearpage

\subsection{Downstream Examples}

\paragraph{MultiRC}

We report examples of downstream evaluation of small models on the MultiRC question-answering task. We report answers of small {\sf Hyena} (153M, trained for $130$B tokens on {\sc The Pile}) and the public checkpoint RWKV-v4 \citep{PENG_RWKV-LM_2021} ($169$M, trained for $332$B tokens on {\sc The Pile}). We select randomized examples with indices being powers of $2$.
Alignment of {\sf Hyena}'s responses to the task format is greatly improved by providing few-shot examples in the prompt, which may be a promising sign for larger models based on the proposed attention-free architecture.
Each example shows responses for specific examples in the validation set (example indices are listed at the beginning of each example).

\begin{tcolorbox}[sharp corners]
    {\tt MultiRC} index: 2, 4 \\ 
    
    \small {\tt \textbf{Passage:} What causes a change in motion? The application of a force. Any time an object changes motion, a force has been applied. In what ways can this happen? Force can cause an object at rest to start moving. Forces can cause objects to speed up or slow down. Forces can cause a moving object to stop. Forces can also cause a change in direction. In short, forces cause changes in motion. The moving object may change its speed, its direction, or both. We know that changes in motion require a force. We know that the size of the force determines the change in motion. How much an object's motion changes when a force is applied depends on two things. It depends on the strength of the force. It also depends on the object's mass. Think about some simple tasks you may regularly do. You may pick up a baseball. This requires only a very small force.} \\ 
    
    {\tt \textbf{Question:} Based on the previous passage, Would the mass of a baseball affect how much force you have to use to pick it up? Is "Less the mass, less the force applied" a correct answer?} \\

    {\tt \textbf{Answer:} Yes} \\ 

    {\tt {\color{red!80} RWKV zero-shot}: The mass of a baseball depends on the force applied. if the mass of the baseball is small, the force applied is small.}  \\ 

    {\tt {\color{blue!80}Hyena zero-shot}: Yes it is.}  \\

    {\tt {\color{red!80} RWKV few-shot}: The mass, less the force applied, is a more important force. it is the force that changes the mass of a baseball.} \\

    {\tt {\color{blue!80}Hyena few-shot}: Yes.} 

    \vspace{1cm}
    
     {\tt \textbf{Question:} Based on the previous passage, What do you apply to an object to make it move or stop? Is "Strength" a correct answer?} \\

    {\tt \textbf{Answer:} No} \\    

    {\tt {\color{red!80} RWKV zero-shot}: The force applied to the object is the force applied to the object.}  \\ 

    {\tt {\color{blue!80}Hyena zero-shot}: Yes.}  \\

    {\tt {\color{red!80} RWKV few-shot}: The force applied to a baseball is a force. it is a force that is applied to a baseball.} \\

    {\tt {\color{blue!80}Hyena few-shot}: Yes.} 
\end{tcolorbox}

\begin{tcolorbox}[sharp corners]
    {\tt MultiRC} index: 32, 64 \\ 
    
    \small {\tt \textbf{Passage:} The film opens with Sunita, a medical student, and her friends working on a project about the human brain. She wants to investigate the curious case of Sanjay Singhania, a notable city businessman, who is reported to have anterograde amnesia.  Her professor denies access to Sanjay's records as it is currently under criminal investigation.  Sunita, nonetheless, decides to investigate the matter herself.  Sanjay is introduced as he brutally murders a man. He takes a Polaroid picture of the man, and writes on it ``done''.  It is revealed that Sanjay has anterograde amnesia where he loses his memory every 15 minutes.  Sanjay uses a system of photographs, notes, and tattoos on his body to recover his memory after each cycle.  It is revealed that Sanjay is ultimately out to avenge the death of his sweetheart Kalpana, and that he is systematically killing the people who were responsible for it.  His main target is ``Ghajini'', a notable social personality in the city. Police Inspector Arjun Yadav, on the case of the serial murders, tracks Sanjay down to his flat and attacks and disables him. Yadav finds two diaries where Sanjay has chronicled the events of 2005 and 2006 . The film flashes back to 2005 as Yadav reads the diary. Sanjay Singhania is shown as the owner of the Air Voice mobile telephone company. In the course of his business, Sanjay sends his men to meet Kalpana, a struggling model, about putting up a billboard above her apartment. The owner of Kalpana's advertising firm misinterprets this as a romantic advance, and in view of a possible lucrative Air Voice ad campaign and other benefits, encourages Kalpana to accept the overture.} \\ 
    
    {\tt \textbf{Question:} Based on the previous passage, Why did Sanjay murder a man? Is "Sanjay" a correct answer?} \\

    {\tt \textbf{Answer:} Yes} \\ 

    {\tt {\color{red!80} RWKV zero-shot}: The answer is "Sanjay".}  \\ 

    {\tt {\color{blue!80}Hyena zero-shot}: The answer is}  \\

    {\tt {\color{red!80} RWKV few-shot}: The question is not about Sanjay's death.} \\

    {\tt {\color{blue!80}Hyena few-shot}: No.}  \\ 

    \vspace{1cm}

      {\tt \textbf{Question:} Based on the previous passage, Every 15 minutes, Sanjay goes through what process, Which frustrates his attempts to avenge the death of his sweetheart? Is "He forgets about facts" a correct answer?} \\

    {\tt \textbf{Answer:} Yes} \\    

    {\tt {\color{red!80} RWKV zero-shot}: Yes, it is a correct answer..}  \\ 

    {\tt {\color{blue!80}Hyena zero-shot}: The answer is.}  \\

    {\tt {\color{red!80} RWKV few-shot}: The answer is "yes".} \\

    {\tt {\color{blue!80}Hyena few-shot}: The answer is "yes.}    
\end{tcolorbox}

\begin{tcolorbox}[sharp corners]
    {\tt MultiRC} index: 128 \\ 
    
    \small {\tt \textbf{Passage:} In 1863, Alexander II re-convened the Diet of Finland and initiated several reforms increasing Finland\'s autonomy from Russia including establishment of its own currency, the markka. Liberation of business led to increased foreign investment and industrial development. Finland also got its first railways, separately established under Finnish administration. Finally, the elevation of Finnish from a language of the common people to a national language equal to Swedish opened opportunities for a larger proportion of the society. Alexander II is still regarded as "The Good Tsar" in Finland. These reforms could be seen as results of a genuine belief that reforms were easier to test in an underpopulated, homogeneous country, than in the whole of Russia. They may also be seen as a reward for the loyalty of its relatively western-oriented population during the Crimean War and during the Polish uprising. Encouraging Finnish nationalism and language can also be seen as an attempt to dilute ties with Sweden.} \\ 
    
    {\tt \textbf{Question:} Based on the previous passage, Alexander II is considered what in Finland since 1863? Is "The good tsar" a correct answer?} \\

    {\tt \textbf{Answer:} Yes} \\ 

    {\tt {\color{red!80} RWKV zero-shot}: The correct answer is "yes".}  \\ 

    {\tt {\color{blue!80}Hyena zero-shot}: Yes, the question is correct.}  \\

    {\tt {\color{red!80} RWKV few-shot}: The Finnish language is a language of the people.} \\

    {\tt {\color{blue!80}Hyena few-shot}: Yes.} 
\end{tcolorbox}

\begin{tcolorbox}[sharp corners]
    {\tt MultiRC} index: 1024 \\ 
    
    \small {\tt \textbf{Passage:} Einstein and Maric married in January 1903. In May 1904, the couple's first son, Hans Albert Einstein, was born in Bern, Switzerland. Their second son, Eduard, was born in Zurich in July 1910. In 1914, the couple separated; Einstein moved to Berlin and his wife remained in Zurich with their sons. They divorced on 14 February 1919, having lived apart for five years. Eduard, whom his father called "Tete" (for petit), had a breakdown at about age 20 and was diagnosed with schizophrenia. His mother cared for him and he was also committed to asylums for several periods, including full-time after her death. The marriage with Maric does not seem to have been very happy. In letters revealed in 2015, Einstein wrote to his early love, Marie Winteler, about his marriage and his still strong feelings for Marie. In 1910 he wrote to her that "I think of you in heartfelt love every spare minute and am so unhappy as only a man can be" while his wife was pregnant with their second child. Einstein spoke about a "misguided love" and a "missed life" regarding his love for Marie. Einstein married Elsa Lowenthal on 2 June 1919, after having had a relationship with her since 1912. She was a first cousin maternally and a second cousin paternally. In 1933, they emigrated to the United States. In 1935, Elsa Einstein was diagnosed with heart and kidney problems; she died in December 1936.} \\ 
    
    {\tt \textbf{Question:} Based on the previous passage, How much time passed, after Albert Einstein's father divorced his mother, that he re-married? Is "3 months and 18 days" a correct answer?} \\

    {\tt \textbf{Answer:} Yes} \\ 

    {\tt {\color{red!80} RWKV zero-shot}: Albert Einstein was born on 3 march 1916 in Gerlin, Germany. he was the son of a German doctor and a german woman.}  \\ 

    {\tt {\color{blue!80}Hyena zero-shot}: "3 months and 18 days"}  \\

    {\tt {\color{red!80} RWKV few-shot}: It is not a correct answer. The exact date is not known.} \\

    {\tt {\color{blue!80}Hyena few-shot}: Yes, according to the previous passage.} 
\end{tcolorbox}

\end{document}